\documentclass[preprint,12pt,authoryear]{elsarticle}

\usepackage{appendix}
\usepackage{amssymb,amsmath}
\usepackage{mathtools}
\usepackage{acronym}
\usepackage{ccg}
\usepackage{makecell}
\usepackage{booktabs}
\usepackage{placeins}
\usepackage{url}
\usepackage[table,xcdraw,HTML]{xcolor}
\usepackage{algorithm}
\usepackage{algorithmicx}
\usepackage{algpseudocode}
\usepackage{lmodern}
\usepackage{latexsym,mathptmx,epsfig,makeidx,letterspace}
\usepackage{tabularx,graphicx}
\usepackage{listings}
\DeclareMathOperator*{\argmax}{argmax}

\definecolor{seencolor}{HTML}{D3FFCE}
\definecolor{unseencolor}{HTML}{FFC0CB}

\begin{document}

\acrodef{mr}[MR]{meaning representation}
\acrodef{ud}[UD]{universal dependency}
\acrodef{ccg}[CCG]{combinatory categorial grammar}
\acrodef{em}[EM]{expectation maximization}
\acrodef{lf}[lf]{logical form}
\acrodef{nlp}[NLP]{natural language processing}
\acrodef{cla}[CLA]{child language acquisition}
\acrodef{cds}[CDS]{child-directed speech}
\acrodef{pos}[POS]{part of speech}
\acrodef{lrd}[LRD]{long-range dependency}
\acrodef{map}[MAP]{maximum a posteriori}
\acrodef{em}[EM]{expectation-maximization}
\acrodef{llm}[LLM]{large language model}
\acrodefplural{lrd}[LRDs]{long-range dependencies}
\renewcommand{\floatpagefraction}{1.0}
\newcommand{\mycommand}[1]{\kern\dimexpr #1 * 4em\relax}

\begin{frontmatter}



\title{Modelling Child Learning and Parsing of Long-range Syntactic Dependencies}

 \author[label1]{Louis Mahon}
 \author[label2]{Mark Johnson}
 \author[label1]{Mark Steedman}
 \affiliation[label1]{organization={School of Informatics, University of Edinburgh},
             country={United Kingdom}}

 \affiliation[label2]{organization={School of Computing, Macquarie University},
             country={Australia}}


%
\begin{abstract}
This work develops a probabilistic \ac{cla} model to learn a range of linguistic phenonmena, most notably long-range syntactic dependencies of the sort found in object wh-questions, among other constructions. The model is trained on a corpus of real child-directed speech, where each utterance is paired with a logical form as a meaning representation. It then learns both word meanings and language-specific syntax simultaneously. After training, the model can deduce the correct parse tree and word meanings for a given utterance-meaning pair, and can infer the meaning if given only the utterance. The successful modelling of long-range dependencies is theoretically important because it exploits aspects of the model that are, in general, trans-context-free.
\end{abstract}

\end{frontmatter}

\section{Introduction}
This paper develops a computational model of \ac{cla}, seeking to understand the process of language acquisition by programming a computer to emulate the learning undergone by the child. We focus specifically on the learning of syntax and semantics, that is, learning the meaning of individual words, and learning the language-specific syntax by which they combine to produce a single meaning for a whole utterance. Unlike other computational approaches to language learning, such as the dominant paradigm of large language models, which make minimal prior assumptions about language structure and attempt to learn everything from the data, we distinguish between what there is reason to believe is innate vs learned. In line with semantic bootstrapping theory \citep{pinker1979formal}, we assume the child possesses a rich system of semantic categories prior to language acquisition, such as actions, individuals and attributes, 
which are combined in sentence meanings or logical forms, and, together with the sentence itself, provide the input to "bootstrapping" the lexical word meanings and syntactic categories of the language in question, via a syntactic and semantic derivation induced by the application of universal rules of functional application and composition.

The theoretical backbone of our model is provided by \ac{ccg} \citep{steedman2000syntactic}. \ac{ccg} is a strongly lexicalized theory of grammar, in which all details that are language-specific, such as the linear order of clausal constituents and their mapping to logical form or meaning, is specified in the lexical entry for categories such as verbs. It is the language-specific lexicon that has to be learned by the child and the computer. The universal \ac{ccg} rules of syntactic combination or merger are assumed to be innate, and exhibit tight-coupling of syntax and semantics, with a one-to-one correspondence between the semantic operations of the logical combinators, and the syntactic operations of combining grammatical constituents. This facilitates learning both in a single unified, computational model. 

Learning takes place when utterances from real child-directed speech taken from the CHILDES corpus \citep{macwhinney1998childes} are paired with a corresponding \ac{lf} representing its meaning\footnote{This refers to the interpretable logical form, rather than the Chomskian 'big L' Logical Form.}
, via semantic annotation such as that provided by \cite{ida2023}. Training uses an expectation-maximization-style algorithm \citep{Neal:99}: the model considers each possible valid analysis, weighted by its current estimated probability, and then increases the estimated probability on all constituents in that analysis in proportion to this weight. (So the high probability of previously learned words contributes to the probability assigned to newly encountered ones.)

The work falls between Universal Grammar (UG)-based approaches that
assume expressive theories of grammar drawn from theoretical
linguistics \citep{Chom:65,Chom:81,Chom:95}, and seek to identify
constraints, such as Freezing Principles and Subset Conditions, that
will make such grammars learnable from paired meaning representations
and strings \citep{Wexler:80,Berw:85,Gibs:94,Fodo:98,Yang:02}, and
Usage-based theories based on memorization of exemplars of child
directed utterance (CDU), with or without subsequent generalization
\citep{Tomasello:03a,Bybee:06,FrankM:07,Bannard:08a,ambridge2020against}.  In comparison with
other UG-based approaches, \ac{ccg} is syntactically of
low, near-context-free, expressive power, and is semantically
surface-compositional, requiring no constraints other than the
universal rules of grammatical composition of ordered adjacent
elements and a universal inventory of lexical types.  In comparison
with Usage-based approaches, our learner allows lexicalization of what
in terms of the adult language are multi-word items, including entire
CDUs, but also embodies a mechanism for automatically decomposing such
items on distributional grounds.

Another prominent model of language acquisition is
\cite{chater2018language}, who characterize it as the learning of a
perceptuo-motor skill. \citeauthor{chater2018language} emphasise that
much relevant information to language learning is forgotten quickly,
necessitating that learning occurs rapidly and in real-time (in this
sense, it is the direct opposite of the radical exemplar theory of
\cite{ambridge2020against}, and in line with our own incremental
approach). Another point they emphasize is the social context in which
the child hears the utterance. We account for the first point by
training on each example only once, one at a time, in the order they
appear to the child. Pragmatic context is not currently represented in
our input to the learner, except insofar as it is implicit in our use
of adjacent CDU meanings as distractors from the intended meaning.

In the area of \ac{nlp}, on the other hand, much work is currently focussed on \acp{llm}, which require too much training data to be plausible models of how humans acquire language. Typically the amount is at least several orders of magnitude more tokens than a human sees in their entire life. Attempts have been made to better learn from a number of tokens more similar to that required by humans \citep{conll-2023-babylm}, but this is more of an \textcolor{black}{engineering challenge to improve sample efficiency to a level consistent with human exposure to language data, rather than an explicit attempt to model} the learning process of a child. Such models still generally employ multi-epoch training, batched parameter updates, and arbitrary text tokenization, which are not plausible features of child language acquisition. Additionally, the datasets and order of presentation are not constrained to be realistic, whereas in our work, we use real \ac{cds}, as suggested, for example, by \cite{dupoux2018cognitive}, and present the utterances once only, in exactly the order that they appear to the child.

Some prior works have aimed to more realistically model child acquisition of syntax and semantics \citep{mahon2024language, abend2017bootstrapping, kwiatkowski2012probabilistic}. Ours extends these in two main respects. Firstly, it widens the set of syntactic constructions that can be handled to the following: intransitives, transitives, ditransitives, modals, progressives, negations, subject inversion questions and, most importantly, long-range dependencies of the sort found in object wh-questions such as ``what do you want?'', and potentially including extraction from embedding. While the model of \cite{abend2017bootstrapping} was equivalent to some probabilistic context-free grammar, in order to handle long-range dependencies with cross-linguistically adequate expressive power, we exploit trans-context-free aspects of \ac{ccg}. (Technically, \ac{ccg} is equivalent to a level 2 multiple context-free grammar (2-MCFG, \cite{Seki:91}.) Secondly, our model is able to produce the entirely correct analysis for sentence-meaning pairs at train time, and even infer the meaning at test time if given only the utterance. \textcolor{black}{While \cite{abend2017bootstrapping} showed some limited ability to infer meanings from the utterance only, our model can do so with a much higher accuracy. Relatedly, \cite{abend2017bootstrapping} did not present evidence of fully correct parse trees for unseen utterances without corresponding \acp{lf}, and it is not clear which of those meanings that were correctly inferred were the result of memorization as single-word utterances. As discussed in Section \ref{subsec:ngram-lexicalisation}, we observe that this is common behaviour for our model early in training. In contrast, we show that our model is able to produce fully correct parse trees for unseen utterances, and therefore can infer corresponding meanings beyond memorization. In Section \ref{subsec:qual-examples}, we present several qualitative examples of such inferred trees across a variety of utterance types.}

The novel contributions of this work are as follows:
\begin{itemize}
    \item modelling the learner`s ability to parse and interpret novel child-directed utterance;
    \item learning a wider variety of syntactic constructions, including object wh-questions, which contain unbounded \acp{lrd};
    \item higher accuracy and robustness across the various measures of learning that we test.
\end{itemize}

\section{Theoretical Underpinnings} \label{sec:theoretical-underpinnings}
Our model deals with syntactic and semantic learning only. It assumes the child either has already learned to segment the speech stream and detect potential word boundaries, as evidenced in even young prelinguistic infants \citep{Matt:99}, or is jointly learning phonotactics and morphology with syntax, as in the model of \cite{GoldbergY:13c}. At that point, the child must learn to combine atomic units (words) to produce a meaning representation that depends on (a) the meaning of the constituent words and (b) the manner in which they combine. Initially, for such a child, both are unknown. 

In our framework, this problem manifests in the following way. When a child hears an utterance “Bambi is home”, we assume that, from a combination of perceptual context and background and innate knowledge, it can approximately identify the meaning of the entire utterance as some object in some state: $home(bambi)$. The task is then to figure out which words correspond to which parts of the meaning representation, and the language-specific principles by which they combine. As well as the correct interpretation, where English subjects precede VPs, others are also possible, e.g. “Bambi” means $home(\cdot)$, “is home” means $bambi$ and subjects follow VPs. 

\subsection{Semantic and Syntactic Bootstrapping}
Semantic bootstrapping 
\citep{pinker1979formal,grimshaw1981form,Brow:73,Bowerman:73b,Schlesinger:71} is a theory arising from the observation that children understand semantic categories, such as action, object or property, prior to learning language, and that these categories help the child learn syntactic categories. For example, \cite{gropen1991affectedness} showed that when children are exposed to a ditransitive verb that means making something move in a certain way, they expect the moving thing to be the direct object syntactically, whereas for verbs that mean making something change its state as a result of something else moving, they expect the moving thing to be the \emph{indirect} object. This shows that knowing the meaning of the words in a sentence can help guide the child to understand the syntax of that sentence. 

Syntactic bootstrapping \citep{gleitman1990structural}, on the other hand, emphasises that prior \emph{syntactic} knowledge guides children`s learning of word meanings (semantic knowledge). For example, the results of \cite{fisher1994better} suggest that, when children are presented with a situation that is ambiguous semantically between two options as to which is the agent, they are able to resolve the ambiguity from their syntactic knowledge as to which noun phrase is the subject and which the object. Specifically, \citeauthor{fisher1994better} presented children with scenes in which a ball was being transferred from an elephant to a rabbit, paired with a sentence containing the nonce word `biffing', and were then asked which familiar word was closest in meaning to `biffing'. If the paired sentence was ``the elephant is biffing the rabbit'', they selected `give' as closest, but if it was ``the rabbit is biffing the elephant'', they selected `receive' as closest, i.e., they made whatever interpretation of ``biffing'' allowed the agent to fall in subject position.

\cite{abend2017bootstrapping} and \citet{mahon2024language} have shown that, for simple transitive sentences, it is possible to leverage both semantic and syntactic bootstrapping, and learn syntax and semantics simultaneously with a single model. 

 In this paper, we extend this method to cover a much wider set of syntactic constructions, including \acp{lrd} of the sort found in wh-questions, of the kinds investigated in CHILDES and other acqusition datasets by \cite{Klima:66a} and \cite{Stro:95}, among others discussed below.
 

\subsection{Combinatory Categorial Grammar.}

We choose \ac{ccg} \citep{steedman2000syntactic} as a theory of grammar suitable
for learning of this sort, because of its tight coupling of syntactic
derivation and semantic composition.

All information that is
specific to a given language, such as English, is specified in \ac{ccg} in the
lexicon, by a syntactic category, such as NP for the proper noun
``Harry'', or $\rm S\bs NP$  for the intransitive verb ``walks''.
The latter category identifies the verbs a {\em function} applying to
constituents of type NP (such as that of ``Harry'') to yield a
sentence (such as ``Harry walks'').   The backward or left-leaning
slash $\bs$ in the category $\rm S\bs NP$ further specifies the
subject NP argument
as having to occur to the left of the verb in this language.

In the present categorial notation,
the convention is that argument-types (such as NP here) always appear
to the right of the slash,  so that the syntactic category of the
English transitive verb ``sees'' is written $\rm S\bs NP/NP$, where
the forward or right-leaning slash $/$ means that the object NP is
found to the right of the verb in this language.   All function
categories are binary or ``curried'', and slashes ``associate to the
left'', so this 
category is equivalent to $\rm (S\bs NP)/NP$, specifying the object as
the first argument to combine.

Each syntactic category is paired with a logical form (lf)
representing its meaning, which in the
case of verbs is also a function, specified as a $\lambda$-term.  For
example, the full lexical entries for the above categories are the
following:
\startxl{ex:lexicon1}
\begin{tabular}[t]{lllllllll}
  ``Harry''&:=&$\rm NP:\it harry$\\
  ``walks''&:=&$\rm S\bs NP:\it \lambda y.walks\,y$\\
  ``sees'' &:=&$\rm (S\bs NP/NP:\it\lambda x.\lambda y.sees\,x\,y$.
\end{tabular}
\stopx
In the case of an SVO language like English, the order of combination
of syntactic subject and object arguments of a trasitive verb like
``sees'' in~\ref{ex:lexicon1} happens to be aligned with the
(object-first) order of combination of the corresponding semantic
arguments $x$ and $y$ at lf.  The latter is assumed to be universal
for transitive predicates across all languages.  However, other
languages are free to align syntactic and semantic combination
differently, as is the case for VSO languages like Scots Gaelic, where
the subject is the first syntactic argument.  Section 4 shows that our
learner allows for this possibility, and considers all possible alignments.

Such categories and their lf meaning representations combine
synchronously via a
number of combinatory rules, of which the two most simple are the
following rules
that respectively apply rightward- and leftward- looking functions like verbs to
their arguments such as noun-phrases:
\startxl{ex:arules}
{\em The application rules:}
\begin{lexlist}
\item Forward application\\
  $\begin{array}[t]{lrcl}
\it X / Y:f&\it Y:a&\Rightarrow&\it X:f\,a
\end{array}$\hfill ($\mathbf{>}$)~~
\item Backward application\\
$\begin{array}[t]{lrcl}
\it Y:a&\it X\bs Y:f&\Rightarrow&\it X:f\,a\end{array}$\hfill ($\mathbf{<}$)~~
\end{lexlist}
\stopx
These rules allow \ac{ccg} derivations like the one shown in Figure 1 for the
simple child-directed transitive sentence ``You lost a shoe'' from the
Adam corpus.

The derivation in Figure 1 uses the application rules only.   However,
the \ac{ccg} lexicon also includes ``type-raised'' categories for NPs,
which have the effect of  exchanging the roles of verbs and NPs as
functions and arguments.   Moreover, as well as the
rules~\ref{ex:arules} of function application, \ac{ccg} also includes
rules of function composition, of which the following is the only
instance used in the present paper:
\startxl{ex:brules}
{\em The composition rules ($\mathbf{B}$)}
\begin{lexlist}
\item Forward composition\\
$\begin{array}[t]{lrcl}
\it X / Y:f & \it Y/Z:g & \Rightarrow & \it X/Z:\lambda z.f(g\,z)
\end{array}$\hfill (${>}\mathbf{B}$)
\end{lexlist}
\stopx
Type-raised categories and composition rules allow some extra
derivations, as shown in Figure 2.  This derivational ambiguity is
harmless, since, as the figure shows, they yield the same logical form
for canonical sentences.  However, they are crucial to the derivation
of long-range dependencies, such as those involved in {\em wh}-questions
like the one shown in Figure 3, which are not otherwise derivable.

Type-raised categories were not included in previous work related
to ours (Abend et al., 2017; Mahon et al., 2024).  In section 4 we
will show that their inclusion, together with that of rules of function
composition, allows our model to learn LRDs of this kind.

\begin{figure}[htb]
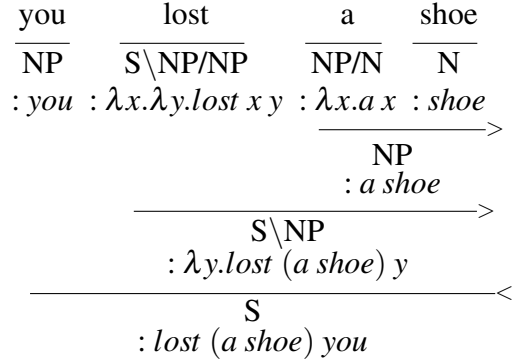

\centering
    
\deriv{4} {
    \text{you}  & \text{lost}          &      \text{a} & \text{shoe}        \\
    \uline{1}    & \uline{1}            & \uline{1}     & \uline{1}         \\
    \text{NP}    & \text{S$\bs$NP/NP}  & \text{NP/N}   & \text{N}          \\[0.25em]
    :you        & :\lambda x.\lambda y. lost\; x\; y         & :\lambda x. a\; x & :shoe                         \\
                 &                      & \fapply{2}                        \\[0.25em]
                 &                      & \mc{2}{\text{NP}}                 \\
                 &                      & \mc{2}{: a\; shoe}          \\
                 & \mc{3}{\hspace{-4em}\hrulefill_{>}}\hspace{-6em} \\

                 & \mc{3}{\text{S\bs NP}}                                   \\
                     & \mc{3}{:\lambda y. lost\; (a\; shoe)\; y}               \\
                 \mc{4}{\hspace{-7em}\hrulefill_{<}}\hspace{-8.5em} \\
                   \mc{4}{\text{S}}                                        \\
                   \mc{4}{: lost\; (a\; shoe)\; you}                                        \\
}
\caption{Example of a \ac{ccg} derivation for a simple transitive sentence from the Adam (English) corpus.} \label{fig:ccg-example}

\end{figure}

\begin{figure}[htb]
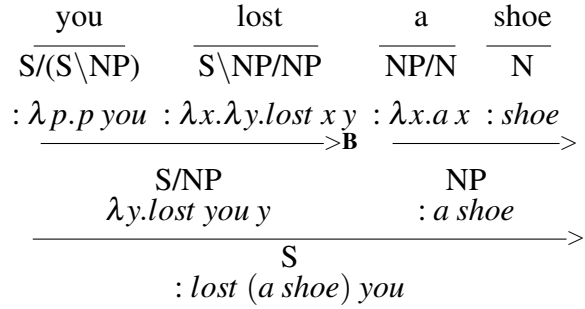

\centering
    
\deriv{4} {
    \text{you}  & \text{lost}          &      \text{a} & \text{shoe}        \\
    \uline{1}    & \uline{1}            & \uline{1}     & \uline{1}         \\
    \text{S/(S$\bs$NP)}    & \text{S$\bs$NP/NP}  & \text{NP/N}   & \text{N}          \\[0.6em]
    :\lambda p. p\; you        & :\lambda x.\lambda y. lost\; x\; y         & :\lambda x. a\; x & :shoe                         \\
                 \fcomp{2}  & \fapply{2}                        \\[0.6em]
                 \mc{2}{\text{S/NP}} & \mc{2}{\text{NP}}                 \\
                 \mc{2}{\lambda y. lost\; you\; y} & \mc{2}{: a\; shoe}          \\
                 \mc{4}{\hspace{-8em}\hrulefill_{>}}\hspace{-9.5em} \\
                   \mc{4}{\text{S}}                                        \\
                   \mc{4}{: lost\; (a\; shoe)\; you}                                        \\
}
\caption{Example of the alternative, type-raise and compose, \ac{ccg} derivation for the sentence in Figure \ref{fig:ccg-example} from the Adam (English) corpus.} \label{fig:left-branch-ccg-example}

\end{figure}

\begin{figure}[htb]
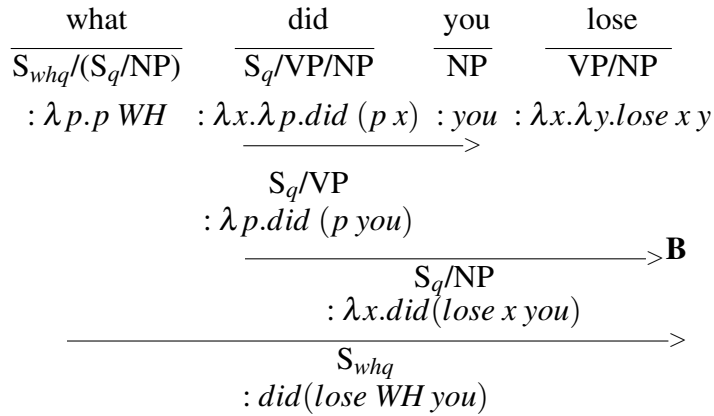

\centering
    
\deriv{4} {
    \text{what}  & \text{did}          &      \text{you} & \text{lose}        \\
    \mc{1}{\hspace{-3em}\hrulefill} \hspace{-3em}    & \uline{1}            & \uline{1}     & \uline{1}         \\
    \text{S$_{whq}$/(S$_q$/NP)}    & \text{S$_q$/VP/NP} & \text{NP} &   \text{VP/NP}  \\[0.6em]
    :\lambda p. p\; \textit{WH}        & :\lambda x.\lambda p. did\; (p\; x)         & : you & : \lambda x.\lambda y.lose\; x\; y \\
    & \fapply{2}  &                         \\[0.6em]
    & \text{S$_q$/VP} &   \\[0.25em]
    & :\lambda p. did\; (p\; you)         & \\
    &  \mc{3}{\hspace{-4.8em}\hrulefill_{>}\mathbf{B}}\hspace{-5.8em}                         \\
    & \mc{3}{\text{S$_q$/NP}} \\
    & \mc{3}{:\lambda x. did(lose\; x\; you)} \\
                 \mc{4}{\hspace{-8em}\hrulefill_{>}}\hspace{-9em} \\
    \mc{4}{\text{S$_{whq}$}}                    \\[0.25em]
                   \mc{4}{: did(lose\; \textit{WH}\; you)}                                        \\
}

\caption{Example of a \ac{ccg} derivation of the object-wh question corresponding to Figure~\ref{fig:ccg-example}.} \label{fig:wh-ccg-example}

\end{figure}

\FloatBarrier

It will be noticed that the logical forms exemplified in Figures~\ref{fig:ccg-example}
through~\ref{fig:wh-ccg-example} are, as a consequence of the process of semi-automatic
annotation of the CHILDES dataset \citep{ida2023}, somewhat
English-specific in comparison to anything we might imagine to be the
form of the universal language of mind to which child language
learners are assumed to have access.  This means that if our learner
were faced with the corresponding French utterances paired with the
same logical form, it would begin by learning a lot of multiword
items, such as ``Qu'est-ce que'' with the meaning of ``what'',
$\it\lambda p.p\;\textit{WH}$, and ``range'' with the meaning of ``put away'',
$\it\lambda x\lambda y.put\,away\;x\;y$.  However, the learner would
still learn from such data, and in many cases generalize to a more
standard lexicon.

\subsection{Long-range Dependencies}
\Ac{lrd} constructions, as we use the term here, are those in which a word depends semantically on a word or set of words that are arbitrarily far away in the sentence, as in ``what did you lose?''. In some grammars (though not in \ac{ccg}), these are treated as filler-gap dependencies, related by a discontinuous operator, such as movement. Figure~\ref{fig:wh-ccg-example} shows the \ac{ccg} derivation of an object wh-question from the Adam corpus. Note the use of the type-raised category on the wh-word `what', and the composition operator on the second-to-last line.\footnote{In wh-questions, such as Figure \ref{fig:wh-ccg-example}, the wh-word is a second-order type. In the full theory, all NPs are type-raised to second-order to capture further coordination/relativization phenomena (see \cite{steedman2000syntactic}), though this is not a part of our model.} The ability to correctly handle \acp{lrd} is essential for accurately modelling real-world language, where such constructions are common. It is also of theoretical importance because it establishes \ac{ccg} as properly including the context-free languages in the Chomsky hierarchy, a property which is known to be necessary to capture natural languages 
\citep{chomsky1957syntactic,shieber1988evidence}.
In our corpora of child-directed speech, object wh-questions appear with high frequency, accounting for 21.6\% of all utterances, and including some of the most common utterances such as ``what do you want?'', ``what are you doing?'', and ``what's that?''. 

\section{Method} \label{sec:method}

\subsection{Probabilistic Model} \label{subsec:prob-model}
The probabilistic model is broadly the same as that described in \cite{mahon2024language}.
In our framework for syntax and semantics learning, each example consists of the string of words in the utterance $w$, the meaning representation $m$, and the parse tree $T$. The parse tree is always unobserved so it is treated as a latent variable. We fit an approximation to the joint data distribution $P(w,m,T)$ via several univariate conditional distributions. 

Typically, \ac{ccg} parsing is discussed in terms of combining constituents via combinatory rules to derive a root. For example, the last step of the derivation in Figure~\ref{fig:ccg-example} uses the Backward Application Rule: $Y, X\bs Y \rightarrow X$. Our learning model, when interpreting a sentence-meaning pair, runs these combinators in reverse, that is, it proceeds by successively splitting a root into smaller chunks until they can be aligned with word spans. We will thus often speak of \ac{ccg} `splits', by which we mean the \ac{ccg} combinators run in reverse. The net effect is that our model considers all possible ways to split up the sentence and the meaning representation so that the semantic units best correspond to the language units. 

For the parse tree, the fit distribution has the form $p_t(y|s)$, where $y$ is either a pair $(s_1, s_2)$ of \ac{ccg} syntactic categories that combine to form $s$, or else a symbol $\it leaf$, indicating that the category should not be split in this parse tree. There is also a distribution $p_r(s)$, that predicts a root category.
The distribution relating word $w$ and meaning representation $m$ has the form $p_w(w|m)$, and similarly for the distribution relating syntactic category $s$ to meaning representation $m$. Following \cite{abend2017bootstrapping}, we first predict a shell \ac{lf}, consisting of semantic types for all non-variables, and then predict the \ac{lf} from the shell \ac{lf}. Thus, for meaning $m$ and category $s$, $p(m|s)$ is decomposed as $p_l(m|e)$ and $p_h(e|s)$. The shell \ac{lf} replaces all non-variable terms with a placeholder marked for the function of the placeholder: verb (for which we write `vconst'), entity, determiner etc. The function is inferred from the CHILDES part-of-speech tag given in the method of \cite{ida2023}. For example the \ac{lf} $\lambda y. lost\, (a\, shoe)\, y$ from Figure~\ref{fig:ccg-example} has shell logical form $\lambda y. \text{vconst}\,y\,(\text{quant}\; \text{noun})$
This allows the model to share representation power for the structure of the logical form across different examples that may have different values for the constants. See \cite{mahon2024language} for details.

Each of these distributions is modelled as a Dirichlet process, to which Bayesian updates are applied at each training example. Taking $p_w$ as an example, the form of the posterior is then
\begin{equation} \label{eq:dp-form}
p_w(w|m) = \frac{n(w,m) + \alpha H(w|m)}{n(m) + \alpha}\,,
\end{equation}
where $n(w,m)$ is the number of times $w$ and $m$ have been observed together in the past, $n(m)$ is the number of times $m$ has been observed in the past, and $H(x)$ is a pre-defined base distribution (see \ref{app:base-distributions}). An analogous definition holds for $p_l$, $p_h$ and $p_t$. 
The alpha parameter is set to $1$ for all distributions, corresponding to a uniform Dirichlet prior across simplices. During training, we set $\alpha=10$ in $p_t$ to encourage exploration of different syntactic structures, and $\alpha=0.25$ in $p_w$ to produce more confident predicted word meanings, which we find helps stabilize syntax learning.

The probability assigned to a full analysis, consisting of a parse tree and a meaning for each leaf node, is the product of the probabilities of all of constituent nodes given their parents. This is a stronger independence assumption than is made in head-dependency models \citep{Coll:97}. Our model would fail to resolve attachment ambiguity such as that between high attachment of ``with''-adjuncts in ``I saw a squirrel with a telescope'', and low attachment, as in ``I saw a squirrel with an acorn''. We expect that our model would handle such ambiguities with the future addition of a "supertagger" \citep{Srin:94, LewisM:14d, Coll:97}. This would be a neural model, e.g. a small encoder-only transformer, which predicts a small set of possible \ac{ccg} categories for each word in the current string context. This model is related to two-factor theories of processing advanced in the psycholinguistic literature by \citet{Ferr:07} and \citet{kahneman2011thinking}, among others. Based on surrounding context, including words like "saw", "squirrel", and "telescope", such a model would learn to predict the category VP$\bs$VP/NP for "with", in contrast to contexts including "saw", "squirrel" and "acorn", which predict N$\bs$N/NP, thus resolving the ambiguity.

\subsection{Training} \label{subsec:training}
\begin{figure}
    \centering
    \includegraphics[width=0.5\linewidth]{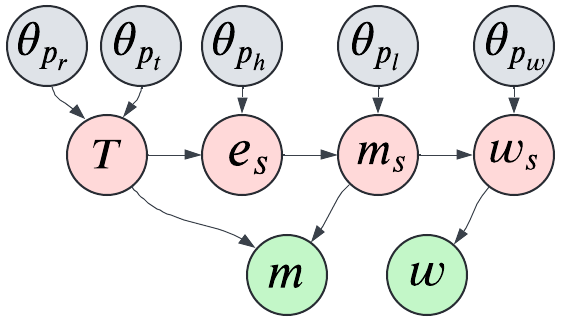}
    \caption{Graphical model for for our probabilistic model. $T$ is the parse tree, $e_s$, $m_s$ and $w_s$ are the leaf-level shell \acp{lf}, \acp{lf} and word, $m$ is the root-level \ac{lf}, $w$ is the utterance and $\theta_x$ is the subset of the full set $\theta$ of model parameters, consisting of the cooccurence counts in the distribution $x$, as described in Section \ref{subsec:prob-model}. Green indicates that a variable is observed, and red indicates unobserved. These colours are for train time, at test time, $m$ would also be red. The fact that $w$ is observed but $w_s$ is not reflects the fact that the model sees the full utterance, but not where the word boundaries should be, and similarly with $m$ and $m_s$.}
    \label{fig:graphical-model}
\end{figure}

The parameter updates described in Section~\ref{subsec:prob-model} require tracking the number of times two different elements co-occur. For example, in $p_w$, the probability of predicting the logical form $\lambda x. \lambda y. lost\,x\,y$ to be realized as the word `lost' depends on the number of times that logical form and word were observed together during training. Because we do not observe parse trees directly, we instead employ an \ac{em} algorithm, as follows. When, at time $t$, the model observes a single training example $(w,m)$, consisting of an utterance as a string $w$ and a corresponding logical form $m$, it uses its current parameter values $\theta^{(t)}$ to estimate a distribution over all possible parses that connect the two. \textcolor{black}{The set of parameters, $\theta^{(t)}$ consists of the occurrence counts in the Dirichlet processes, e.g. the number of times a leaf meaning such as $\lambda x.\lambda y.lost\; x\; y$ occurs with a word such as `lost'. The set of parameters can grow throughout training as new occurrences are observed.} The probability assigned to a parse tree $T$ and the training example $(w,m)$ is the following product
\begin{gather} \label{eq:p-def}
p(w,m,T|\theta^{(t)}) = p_r(r)\prod_{s'} p_t(s_1, s_2|s') \prod_{s} 
p_t(\text{leaf}|s)p_h(e_s|s)p_l(m_s|e_s)p_w(w_s|m_s)\,,
\end{gather}
where $r$ is the root category, $s'$ ranges over all non-leaf nodes in $T$, $s_1$ and $s_2$ are the children of $s'$, $s$ ranges over all leaf nodes in $T$, and $e_s$, $m_s$ and $w_s$ are, respectively, the shell logical form, the logical form, and the word aligned to $s$ in $T$. Probabilities for the leaf-level \acp{lf} $m_s$ are determined by the root \ac{lf} $m$, together with the shell \acp{lf} of each node, $e_s$, and the model parameters $\theta$, and the same is true for $w_s$ and $w$. 
As $(w,m)$ is something we observe, we are interested in the conditional probability of a given parse tree
\begin{equation} \label{eq:bayes-posterior}
p(T|w,m,\theta^{(t)}) = \frac{p(w,m,T|\theta^{(t)})}{\sum_{T' \in \mathcal{T}} p(w,m,T'|\theta^{(t)})}\,,
\end{equation}
where $\mathcal{T}$ is the set of all allowable parses of $(w,y)$.
For each parse tree, the co-occurrences that it gives rise to are recorded in proportion to the parse tree's probability. Combining the standard \ac{em} update rule with the Bayesian update for the Dirichlet process, then, for each parameter in $\theta_i \in \theta$ that tracks the co-occurrence of two elements $a$ and $b$, the update rule is given by
\[
\theta_i^{(t+1)} = \theta_i^{(t)} + \mathbb{E}_{T \sim p(T|w,y,\theta^{(t)})}[\delta_T(a,b)]\,,
\]
where $\delta_T(a,b)$ is an indicator function that is 1 if $a$ and $b$ co-occur in $T$ and 0 otherwise. \textcolor{black}{The relation between the variables $T$, $e$, $m$, $w$ and $\theta$ is given in Figure \ref{fig:graphical-model}.}

The set $\mathcal{T}$ of allowable parse trees is the set of all valid \ac{ccg} parse trees that have the observed \ac{lf} as root, the words in the observed utterance as leaves, and that have congruent syntactic and semantic types.\footnote{We use `semantic/syntactic type' and `semantic/syntactic category' interchangeably.} We require the semantic category to be congruent with the \ac{ccg} category, for each node. \ac{ccg}'s tight coupling of syntax and semantics provides a straightforward mapping from syntactic to semantic categories. In particular, the \ac{ccg} atomic categories $S$ and $NP$ correspond to the Montagovian $t$ and $e$ respectively, and the slashes in non-atomic categories correspond to functions between types.\footnote{In the full theory, NP is treated as a schema, and type-raised just in time during parsing to an appropriate form as determined by the context. We pass over this detail here and simply treat NP as a category that can combine directly with others.} For example, the \ac{ccg} transitive verb category $S{\bs}NP/NP$ has the semantic type ${<}e, {<}e,t{>}{>}$. See \cite{steedman2000syntactic} for further details. If, when expanding a parse tree, any node violates these constraints, then that branch of search is terminated.




This constraint is based on the assumption that the child knows the semantic type of a \ac{lf} (or fragment thereof on some internal parse node).  For example, in the derivation of `you lost a shoe', from Figure \ref{fig:ccg-example}, the child knows that the constituent S\bs NP: $\lambda y. lost\; (a\; shoe)\; y$, in the second-to-last row, has semantic type ${<}e,t{>}$.
These constraints 
speed up training significantly. 
and make training more robust by removing the noise of updates from inconsistent parse trees\textcolor{black}{, i.e., those that are not in $\mathcal{T}$}. We believe this is the reason for our improved robustness to noise in the \acp{lf} over \cite{abend2017bootstrapping}, as described in Section~\ref{subsec:distractor-results}.

The computation of all allowable parse trees can be performed efficiently by caching the probability of each subtree.

\begin{figure}
    \centering
    $\colorbox{seencolor}{not\; (can\; (see\; (the\; music)\; you))}$
    \includegraphics[width=\textwidth]{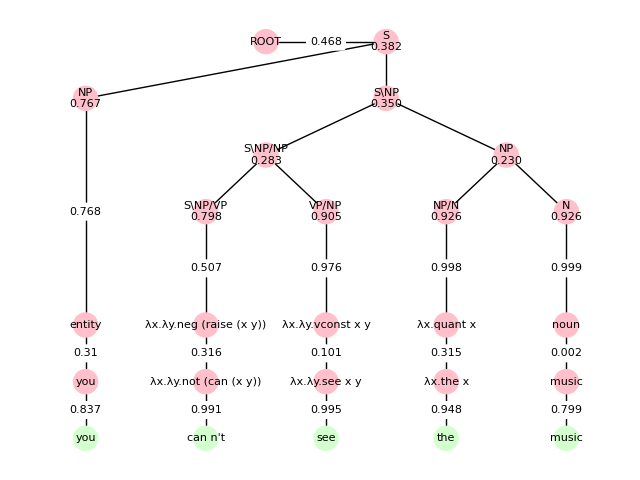}
    \caption{One of the parses considered by the learner for this example.  Given information is in \colorbox{seencolor}{green}, inferred information is in \colorbox{unseencolor}{pink}. As this is train time, the model sees both the utterance and the root \ac{lf}. Strictly speaking, the model only the full utterance and does not see individual words because the boundaries between words are not given. This is reflected in Figure \ref{fig:graphical-model}.}
    \label{fig:example-parse-tree}
\end{figure}

\subsection{Worked Example}
Here we present a worked example for a single training example. Recall 
that each training example consists of an utterance and corresponding logical form, and the learner considers the set $\mathcal{T}$ of all compatible parses, i.e. all parses with the observed \ac{lf} as root, the observed utterance as leaves and that obeys the constraints described in Section \ref{subsec:prob-model}. We describe the training updates for a single, correct parse for the example ``you can't see the music'', as shown in Figure \ref{fig:example-parse-tree}. The prediction of the tree proceeds from the root. We will detail the predictions made along the left-most branch from the root to the leaf `you', with all other predictions being made similarly. All predictions are made using the Dirichlet processes described in Section \ref{subsec:prob-model}. 

First, the learner uses the Dirichlet process, $p_r$, to predict a possible root category, here $S$ with probability $0.548$. Next, it uses the Dirichlet process for syntactic prediction, $p_t$, to predict a probability for all possible splits of this root syntactic category $S$. The split shown here is into NP and S\bs NP, with probability $0.395$. Then, for the left child node NP, it again uses $p_t$ to predict a probability for all splits into further child syntactic categories, or alternatively that this node is a leaf. Here, the prediction is that the node is a leaf, which $p_t$ gives probability $0.765$. This indicates that a substantial portion of the NP nodes it has observed in the past have been leaf nodes. The bulk of the remaining probability mass is taken up by the split into (NP/N, N). 

This ends the syntactic stage of prediction, and the role of $p_t$. The task now is to predict what meaning and word(s) should correspond to this NP leaf node. To this end, it first uses $p_l$ to predict a probability for all possible shell \acp{lf}, the one of which shown here is ${<}e{>}$, with probability $0.75$. Again, the bulk of the remaining $0.25$ probability mass is taken up by the possibility a bigram of determiner plus noun, with shell \ac{lf} $quant\; noun$. Interpretations of this sort are discussed in Section \ref{subsec:ngram-lexicalisation}. The fact that the input, or conditioning variable, for $p_l$ is the syntactic leaf only, and not any other information from the tree, is the manifestation of the independence assumption discussed in Section \ref{subsec:prob-model}. 

Next, the learner uses $p_m$ to predict probabilities for all likely meanings of the ${<}e{>}$, here $you$ with probability $0.341$. This stage of predicting \ac{lf} given shell \ac{lf} generally gives the smallest probability of all predictions on the tree, because there are many different meanings corresponding to a given semantic type. The relatively high value of $0.341$ reflects the high frequency of $you$ as a meaning in the dataset. 

Finally, $p_w$ predicts probabilities for all possible words that could correspond to the meaning $you$. The probability of $0.896$ thus represents the learner`s belief, at this stage in training that the word `you' is a realisation of the meaning $you$. The remaining $0.104$ is made up of a long tail of other incorrect meanings, arising from various incorrect interpretations of previous training examples. We observe that this figure continues to reduce to about $0.05$ by the end of training.

The total probability for this tree and leaves given the root \ac{lf} is then computed by multiplying all predicted probabilities at all locations on the tree, which gives $\sim 2.736e-20$.
Given that we observe the leaves, we condition on this event by diving by the sum of the probabilities of all elements of $\mathcal{T}$. Here, that sum turns out to be $7.079e-20$, so the conditional probability for the tree in Figure \ref{fig:example-parse-tree} is
\[
\frac{2.736e-20}{7.079e-7} \approx 0.386\,.
\]
Thus, we assume that this tree is `observed' weighted by this probability, and the counts for the pairings in this tree are updated in the corresponding Dirichlet processes are updated by $0.386$. 

This same procedure is repeated for every element of $\mathcal{T}$. In practice, for the corpora we use, this is generally $50-100$ trees.

\subsection{The Learner as a Parsing Model} \label{subsec:parsing}
After training, it is possible for the learner to parse novel utterances to infer their syntax trees. \textcolor{black}{That is, the learner observes only an utterance, no corresponding \ac{lf}, as well as the parse tree. This differs from train time, where it observes the utterance and the \ac{lf}, and only the parse tree is unobserved. Formally, on observing $w$, we seek $\argmax_{m, T} p(w, m, T |\theta_{final})$, where $p(\cdot)$ is as in \eqref{eq:p-def}, and $\theta_{final}$ are the model parameters after training. Computing this exactly is intractable, so we approximate} using a combination of beam search and a Cocke-Young-Kasami (CYK) based chart-parsing algorithm for \ac{ccg}. First, we marginalise the Dirichlet distributions $p_l$ and $p_h$. This is done as follows, using $p_l$ as an example
\begin{equation} \label{eq:marg-formula}
p_l(x) = \frac{\sum_{v_{sh} \in V_{sh}} p_l(x|v_{sh}) \sum_{v_m \in V_m} c_l(v_{sh}, v_m)}{\sum_{v_{sh} \in V_{sh}} \sum_{v_m \in V_m} c_l(v_{sh}, v_m)}\,,
\end{equation}
where $c_l(x,y)$ is the raw count from the Dirichlet process of the occurrence of shell meaning $x$ with meaning $y$. The denominator in \eqref{eq:marg-formula} thus expresses the number of times any pair of meaning and shell meaning have been observed by the learner. 

These marginal distributions then facilitate a beam search to predict a beam of highest probability \ac{lf}-category pairs for each word span in the utterance. Recall that the learner also considers interpretations in which multiple words form a single lexical item, so this beam search is run on all contiguous spans in the utterance. Then we continue the beam search into CYK-based chart parsing to predict a \ac{ccg} syntax tree. The full method for beam search of leaf nodes is specified in Algorithm \ref{alg:method}. After this, we run CYK for \ac{ccg} to predict a parse tree for the entire utterance.

\begin{algorithm}
\caption{ \small  \small Algorithm for parsing unseen utterances to infer the root lf.} \label{alg:method}
\begin{algorithmic}
    \State $V_m \gets$ vocabulary of all observed \acp{lf}
    \State $V_{sh} \gets$ vocabulary of all observed shell-\acp{lf}
    \Function{SearchLeafSpan}{ws}
    \State $B \gets []$
    \For{$v_m \in V_m$}
        \State $p \gets p_w(ws|v_m)p_m(v_m)$
        \State append $(v_m, p)$ to $B$
    \EndFor
    $B \gets$ top 10 entries in $B$, ranked by $p$
    \For{$(v_m,p) \in B$}
        \For{$v_t \in V_t$}
            \State $p' \gets p_m(v_m|v_t)$
            \State $p_{leaf} \gets p_t(\text{leaf}|v_t)$
            \State $p \gets pp'p_{leaf}\frac{p_t(v_t)}{p_m(v_m)}$
            \State append $(v_t, v_m, p)$ to $B$
        \EndFor
    \EndFor
    \State $B \gets$ top 10 entries in $B$, ranked by $p$
    \State \Return $B$
    \EndFunction
\end{algorithmic}
\end{algorithm}


Note that a typical parsing model is given the meaning and possible categories of each word, and then learns to select the correct categories from amongst these possibilities and to form the syntax tree. Our learner, on the other hand, learns the meaning and categories of the leaves from scratch, as well as learning to form the syntax tree. Due to \ac{ccg}'s close coupling of syntax and semantics, the parse tree, along with a meaning for each leaf, then allows us to compute the meaning for the entire utterance. This is used as an evaluation method in Section \ref{subsec:meaning-results} below.

\section{Results} \label{sec:results}

\subsection{Data} \label{subsec:data}
The data we use for training and testing is taken from Brown's \citeyear{Brow:73} Adam corpus, containing transcribed child-directed speech in North American English to a child ranging in age from 2 years 3 months to 3 years 11 months. It consists of 9314 tokens and 5320 utterances, which amounts only about 2\% of the data used by the child for language acquisition in the relevant period \citep{gilkerson2017mapping}. However, the child is simultaneously learning other skills, such as social, perceptual and motor skills, whereas our model isolates the problem of learning syntax and word-level semantics.

The utterances are extended with \acp{lf}, specifically lambda calculus expressions, as in Figures~\ref{fig:ccg-example}, \ref{fig:left-branch-ccg-example} and~\ref{fig:wh-ccg-example}. Each training example is then a pair of an utterance, as a string in English, and a corresponding \ac{lf}. The \acp{lf} are produced using the method of \citet{ida2023}, which first forms a \ac{ud} parse of the utterance and then uses the UDepLambda library \url{https://github.com/sivareddyg/UDepLambda}, to convert these parses into \acp{lf}. 
The \ac{ud} parses were automatically checked for correctness using the checker at \url{https://github.com/UniversalDependencies/tools/}. The tokenization is taken from the CHILDES corpora. 

\subsection{Word-order Learning} \label{subsec:word-order}
\begin{figure}
    \centering
    \includegraphics[width=0.8\linewidth]{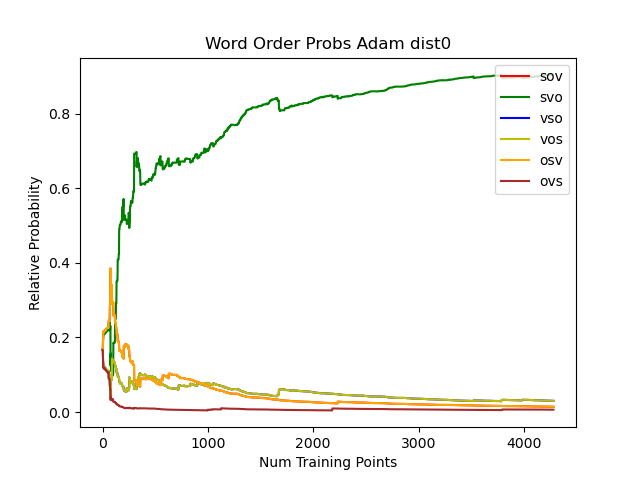}
    \caption{Relative word order probabilities, over the course of training, for each of the six possible word orders for S, V, and O as reflected by verb category. \textcolor{black}{SVO order is learnt confidently within the first 500 examples, and rises to 90\% by the end of training.}}
    \label{fig:word-order-probs-dist0}
\end{figure}

Prior works on similar models to ours \citep{abend2017bootstrapping, mahon2024language}, evaluated the learning of word order by examining the model`s internal parameters to calculate the prior probability, that is, the probability before observing any utterance or \ac{lf}, of the transitive verb category $S\bs NP/NP$. Specifically, this is the sum of all parse trees that would yield this category, namely the right-branching derivation consisting of forward application and backward application, as shown in Figure \ref{fig:ccg-example}, and the left-branching derivation consisting of forward application and forward composition, as shown in Figure \ref{fig:left-branch-ccg-example}. 
Figure \ref{fig:word-order-probs-dist0} shows these relative probability scores over the course of training. Clearly, the prior expectation for SVO order is learnt rapidly and confidently, which reproduces the results for similar models in \cite{abend2017bootstrapping, mahon2024language}. 

\subsection{Meaning and Category for Individual Words} \label{subsec:single-word-results}
\begin{figure}
    \centering
    \includegraphics[width=0.8\linewidth]{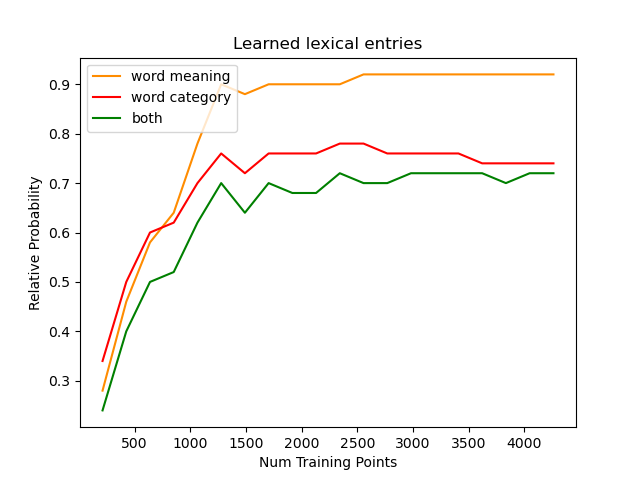}
    \caption{Learning curves for meanings (\acp{lf}) and syntactic categories for the 50 most common words in the corpus (see \ref{app:annotated-lexicon}). Both are learnt successfully, with word meaning higher than word category.}
    \label{fig:words-learning}
\end{figure}

We also follow \cite{mahon2024language} and evaluate the learned meanings and syntactic categories for individual words. Using Bayes` rule, we can obtain a prediction for the meaning and category of a given word. \textcolor{black}{Specifically, the inferred meaning $m'$ for a word $w$ is given by }
\begin{align}
    m' \approx \argmax_m p_w(w|m)p_w(m) = \argmax_m p_w(w,m)\,, \label{eq:lf-pred}
\end{align}
\textcolor{black}{where the last quantity is approximated by the observed number of times that $w$ and $m$ co-occur in the Dirichlet process, which is essentially the numerator in the DP } For example, for the modal word `can', the learner should predict that the meaning is the ``raising to subject'' verb $\lambda p.\lambda y.can\; (p\; y)$\footnote{This is the `can' of ability, rather than of possibility.}. The predicted category is calculated analogously and in this case is S$\bs$NP/VP. We do this for the 50 most common words, and manually evaluate whether they are correct (our annotations are shown in \ref{app:annotated-lexicon}). Unlike \cite{mahon2024language}, who report only a single figure for accuracy after training, we evaluate this throughout training, to produce a learning curve, which is shown in Figure \ref{fig:words-learning}.

The model learns meanings and categories for these words to a similar degree to \cite{mahon2024language}: 90+\% for meaning and $\sim 70\%$ for category. These are learnt quickly in the first 1000 utterances. The learning of category then plateaus and fails to learn the final 30\% of cases. This is due to the learner not having a systematic representation of person, tense and number, which leads to it often confusing the categories S$\bs$NP, which is an inflected verb phrase, and VP, which is an infinitival verb phrase. For example, for the word `are' in the context of expressing identity, such as `those are yours', meaning $equals\; yours\; those$, it predicts the category VP/NP, when it should predict S$\bs$NP. Recall from Section \ref{sec:method}, that the model predicts the shell \ac{lf} from the syntactic category, and so, after applying Bayes` rule, the experiments here predict the category from the shell \ac{lf}, but currently, the shell \ac{lf} is the same for inflected and infinitival verbs. This is discussed further in Section~\ref{sec:discussion}.

\subsection{Understanding Full Utterances} \label{subsec:meaning-results}
Going beyond merely showing the relative, general preference for SVO order, in this work we examine the learner`s ability to analyze entire utterances correctly. We do this in two different ways. In the first, called `select acc' below, we present the model with a single utterance and a set of 5 \acp{lf}, only one of which is correct, and for each of these \acp{lf}, measure the estimated probability of observing the utterance paired with that \ac{lf}. The incorrect \acp{lf} are selected from immediately before and after the utterance as it appears in the corpus, similar to the distractor setting (discussed in Section \ref{subsec:distractor-results}), with $n=4$. We mark an example correct if and only if the probability is highest for the correct \ac{lf}. In the second, we present only the utterance, and the model must infer the \ac{lf} using the method described in Section \ref{subsec:parsing}, which is marked correct only if it exactly matches the true \ac{lf}. In this setting, the model may encounter utterances which include words that have not been seen before in training, i.e. words that are new to the child. \textcolor{black}{In Section \ref{subsec:fast-mapping}, we showed that the model can learn meanings of novel words from syntactic knowledge alone, i.e. perform syntactic bootstrapping. However, this test setting presents the model with the utterance only, not the \ac{lf}, so the model has no access to the meaning of the novel word and is prevented from making the correct interpretation of the utterance. One may, therefore, prefer to exclude these utterances from testing.} We present results from both settings, one which includes these utterances with unseen words, which are all then scored as incorrect, and one in which they are excluded. 

\begin{figure}
    \centering
    \includegraphics[width=0.8\linewidth]{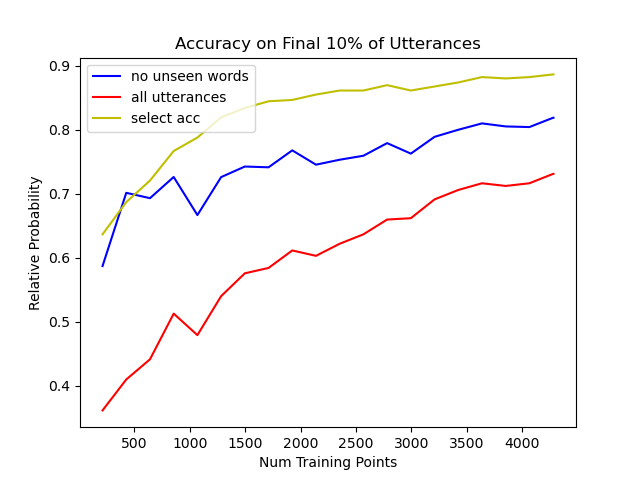}
    \caption{Ability of the learner to infer the meaning of a novel utterance: `select acc' is the fraction of time it gives the highest probabilty to the correct \ac{lf} from a set of candidate \acp{lf}; `all utterances' and `no unseen words' are the fraction of utterances for which the model infers the correct \ac{lf} from the utterance alone, measured, respectively, on all utterances in the test set, and all utterances in the test set that do not have previously unseen words. The model accuracy improves steadily and reaches a high final level by all three measures.}
    \label{fig:test-lfs}
\end{figure}

These three different measures of accuracy are computed, throughout the course of training, on the final 10\% of utterances in the corpus, which we use as a held-out test set. All three show a steady increase, and have not yet plateaued at the end of training, suggesting they would continue to improve if given more training data. The red line, which shows the accuracy on all test items, including those with unseen words, is of course always lower than the blue line, where these test items have been removed. The yellow line is generally the highest, reaching 88\% by the end of training, though at least one point during training, it is exceeded by the accuracy with novel words excluded (blue line). 

The test set for the blue line is changing slightly over training, specifically the number of points being excluded is decreasing as the set of words seen by the model increases. Therefore we suggest the final point reached by the blue line ($\sim 80\%$) is more revealing of the ability of the model after training, but the trajectory of the red line is more revealing of the course of learning.

Note that utterances with unseen words can be, and indeed often are, correct by the first evaluation method (green line). By the end of training, there are still 53 utterances, out of 476 in the test set, with novel words. The yellow line has a final error rate of only $11.5\%$, and gets the correct answer for 27 out of these 53 utterances with novel words. This shows that the model can still make reasonable interpretations even in the presence of a novel word: if it has some rough idea of what the meaning for the entire utterance might be, it is still often able to deduce the correct analysis. This property of the model is examined further in Section \ref{subsec:fast-mapping}.

\subsection{Accuracy by Construction Type} \label{subsec:acc-by-const-type}

\begin{figure}
    \centering
    \includegraphics[width=0.85\linewidth]{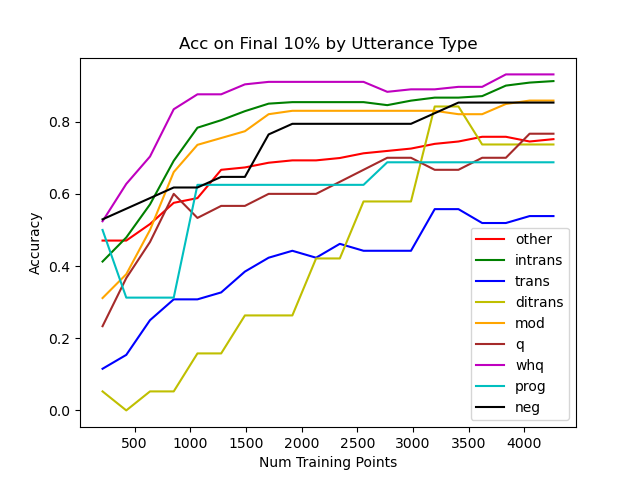}
    \caption{Breakdown of inferred meaning accuracy by different construction types, using the `all utterances' measure. This shows the model is able to infer correct root meanings for a variety of construction types, and the high average accuracy from Figure \ref{fig:test-lfs} is not just the result of a few construction types.}
    \label{fig:parse-acc-by-type}
\end{figure}

Figure \ref{fig:parse-acc-by-type} shows the accuracy of the utterance meanings, separated by the construction type of the utterance. This includes all utterances, even those with novel words. Utterances that exhibit multiple listed features are counted in all the corresponding categories. For example, a negated modal like ``you can't see it'', is counted under `modal' and 'neg'. The measure of accuracy used is the accuracy of the inferred root \acp{lf} when presented with an unseen utterance, i.e., the red line from Figure \ref{fig:test-lfs}. This accuracy increases steadily through training for all utterance types. This shows that the high average accuracy from Figure \ref{fig:test-lfs} is not restricted to just a few types of syntactic construction. Rather, our model learns to infer the correct parse with high accuracy for a variety of syntactic constructions.

The curves in Figure \ref{fig:parse-acc-by-type} are not inter-comparable, and in particular should not be taken as indicating order of acquisition, because the groups are quite different in size and diversity. For example, transitives (`trans', dark blue line) show the lowest final accuracy, but this reflects the fact that the set of simple transitive sentences in our data is largest and most diverse of those presented. Similarly, the fact that wh-questions (`whq', magenta line), show one of the highest accuracies is largely due to that construction type being less frequent and occupied to a greater extent by a few commonly occurring examples, such as `what are you doing?' and `what's that?'. There are still several examples of less common wh-questions in our test set, and we show the full predicted analyses for some of these in Section~\ref{subsec:qual-examples}.

\textcolor{black}{Note that this relationship between low variability and higher accuracy refers to variability of the test items, and is not counter to the evidence that high input variability improves language learning in children (\cite{Huttenlocher:10a} and references therein), which refers to variability in the train items. The low variability in wh-questions in the dataset means the model is only tested on a small set of utterances, whereas for transitives, there is a much higher diversity of utterances it is tested on.}


\subsection{Accuracy on Whq Words' Categories} \label{subsec:wh-results}
Because the ability to model \acp{lrd} of the sort found in object wh-questions is one of the contributions of our model, we present a further experiment focussing specifically on the accuracy for whq words. Figure \ref{fig:parse-acc-by-type} already showed that the accuracy in inferring root \acp{lf} for novel utterances is high for wh questions. However, as we noted in the preceding section, that result does not necessarily reflect a high accuracy in the predicted syntactic categories for wh questions, because it is possible for the model to choose an incorrect or at least non-standard syntactic analysis which nevertheless produces the correct \ac{lf}. Some examples involving lexicalization of multi-world expressions are discussed further in Sections \ref{subsec:qual-examples} and \ref{subsec:ngram-lexicalisation}. Here, we show the accuracy for the syntactic category of the wh-word as it appears in fronted wh-questions in our test set, relative to ground truth categories that we annotated manually. We do not report separate scores for subject and object categories because the nature of the CHILDES data is that there are too few subject questions for the model to learn them effectively. 

\begin{figure}
    \centering
    \includegraphics[width=0.85\linewidth]{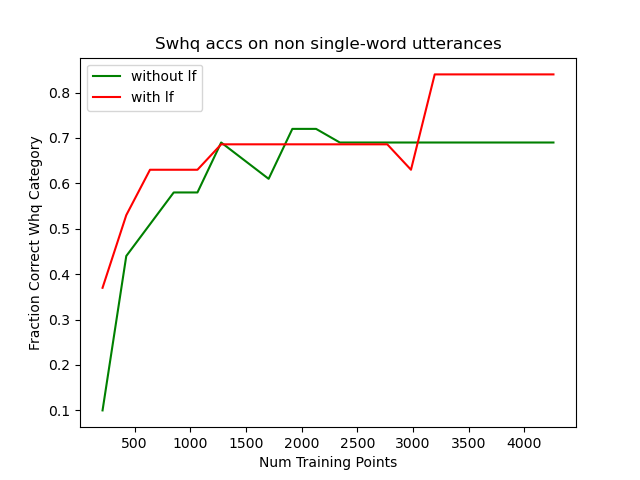}
    \caption{Accuracy of the assigned syntactic categories for whq words in our test set, relative to ground truth categories that we annotated manually. The red line corresponds to the traintime setting where the learner sees both the utterance and the \ac{lf}, the green line to the testtime setting where it sees only the utterance. In both settings, the model is scored as correct whenever its favoured analysis contains a leaf with the correct whq word, meaning and category, and incorrect otherwise. \textcolor{black}{The model reaches a high score by both measures, which shows that it is able to correctly assign syntactic categories to wh words at the leaf level.}}
    \label{fig:whw-cat-acc}
\end{figure}

The learning curves are shown in Figure~\ref{fig:whw-cat-acc}, for the settings where the lf is seen (red line) and unseen (green line) settings. For both settings, the accuracy increases through training and reaches a high accuracy at the end of training (about 85\% with the \ac{lf} and 70\% without). Although we do not distinguish between subject and object questions, we note that the set of wh-questions in the dataset consists almost entirely of object questions. This result is broadly consistent with observations of relatively early acquisition of object questions in children \citep{Stro:95, de1990acquisition, Klima:66a}. It shows that the learner not only learns to infer the correct root \ac{lf} for long-range dependencies of the sort found in object wh-questions, but also learns to model the syntax of these utterances by giving the correct syntactic category to the wh-word, and therefore also to the rest of the sentence.

\subsection{Distractor Settings} \label{subsec:distractor-results}
Our training setting, as described in Section \ref{subsec:training}, presents the learner with a single \ac{lf} for each utterance, i.e., it is told the single correct meaning for the corresponding utterance. In the case of human learning, however, it is more realistic to assume that the child apprehends several possible meanings when it hears an utterance, and does not know, \textit{a priori}, which of these possible meanings the utterance expresses. To simulate this uncertainty, we repeat the experiments from Section \ref{subsec:word-order} and \ref{subsec:meaning-results} with varying numbers of `distractor' \acp{lf} presented alongside the true \ac{lf}. The learner is then free to consider any of these \ac{lf}s as the meaning of the utterance. 
When there is a single tree that the model is very confident in, then the probability from this tree dominates anyway, and overall there is little effect from the distractor trees. However, when there is no such single confident interpretation, the distractor trees significantly reduce the probability on the trees from the correct \ac{lf}, including the correct tree, \textcolor{black}{and add probability, and hence parameter updates, corresponding to the incorrect trees from the distractor \acp{lf}. If the learning trajectory is not stable, the updates from these incorrect trees can derail the model.}

In the real child learner, the distractor logical forms presumably originate in the child's perception and understanding of the state of the world and the conversation, which our model does not directly represent. In our experiment, we take as a proxy for such distractors, the logical forms from the utterances immediately following and preceding the given utterance. Specifically, the $n$ distractor setting takes the $\lfloor n/2 \rfloor$ previous examples and the $\lceil n/2 \rceil$ following examples.

For example, in Adam, training examples 226-228 are as follows:

Data point 226:
“you blow it”--$blow\; you\; it$

Data point 227:
“you can blow”--$can\; (blow\; you)$

Data point 228:
“you do it”--$do\; you\; it$

Thus, in the two distractor setting, when training on training example 227, we include the parse trees from all three of these \acp{lf}. In this case, one possible interpretation takes the \ac{lf} from training example 226–blow(you,it) $blow\; it\; you$--and interprets “you” as meaning $you$, “can” as meaning $it$, “blow” as meaning $\lambda x.\lambda y.blow\; x\; y$, and the sentence as being in SOV order. 

\begin{figure}
    \centering
    \includegraphics[width=0.45\linewidth]{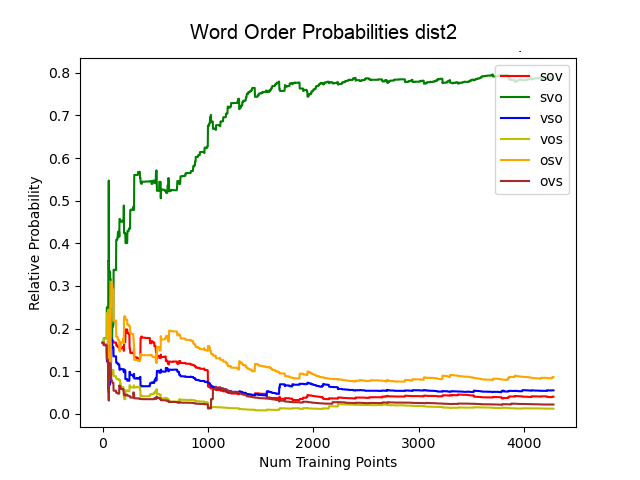}
    \includegraphics[width=0.45\linewidth]{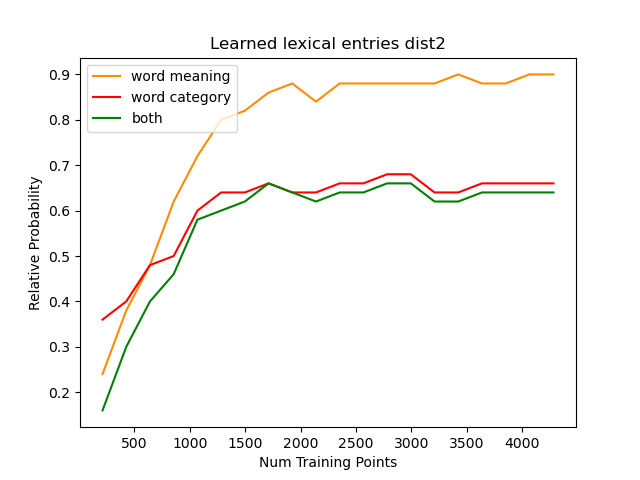}
    \includegraphics[width=0.45\linewidth]{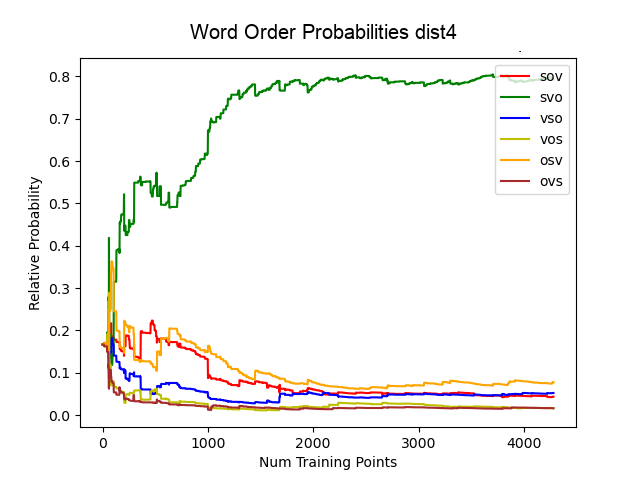}
    \includegraphics[width=0.45\linewidth]{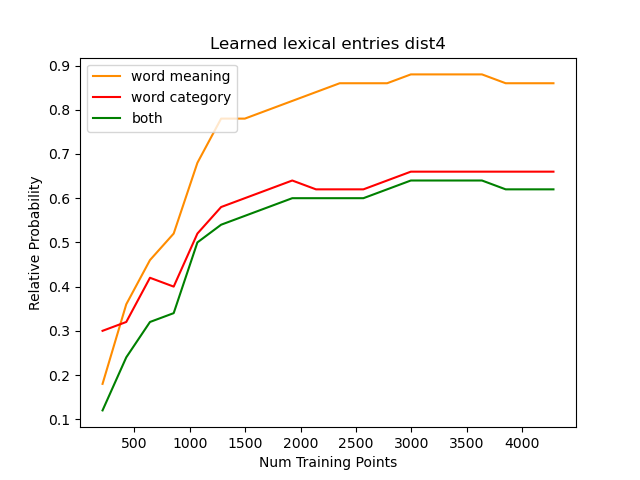}
    \includegraphics[width=0.45\linewidth]{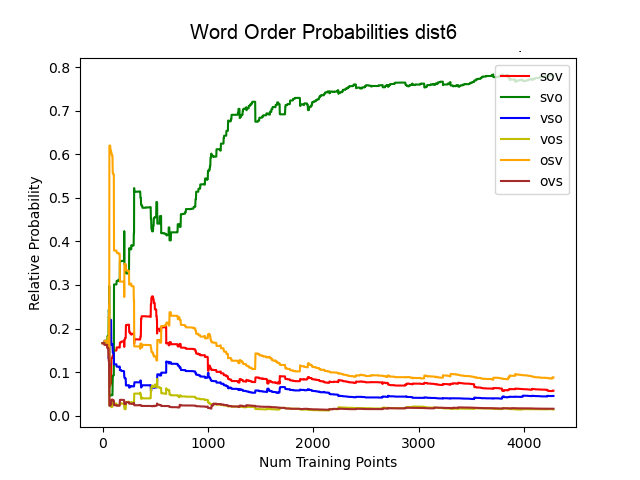}
    \includegraphics[width=0.45\linewidth]{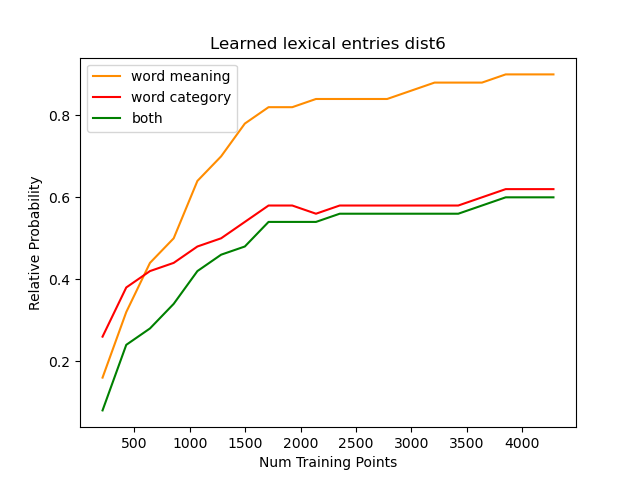}
    \includegraphics[width=0.45\linewidth]{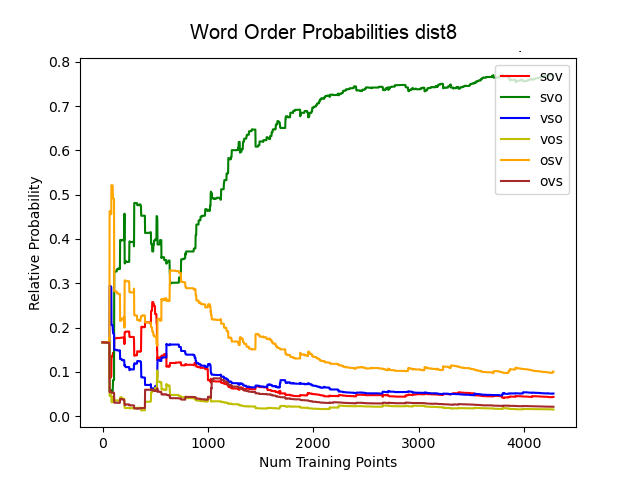}
    \includegraphics[width=0.45\linewidth]{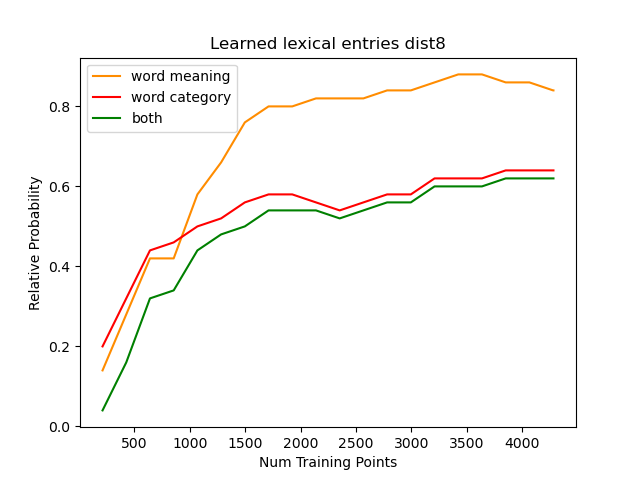}
    \caption{Repeats of experiments for word order of S, V, and O as reflected by
transitive category, and word meaning/category learning, with increasing numbers of distractor \ac{lf}: 2 in the top row, 4 in the second row, 6 in the third row and 8 in the fourth row. Cf. Figure \ref{fig:word-order-probs-dist0}. \textcolor{black}{In this plot, and throughout the paper, a plot title ending in `distN' indicates that there were N distractors present. This shows the learning of word order and word meanings and categories is robust to the presence of distractor meanings during training.}}
    \label{fig:word-order-lex-entries-distractors}
\end{figure}

\begin{figure}
    \centering
    \includegraphics[width=0.45\linewidth]{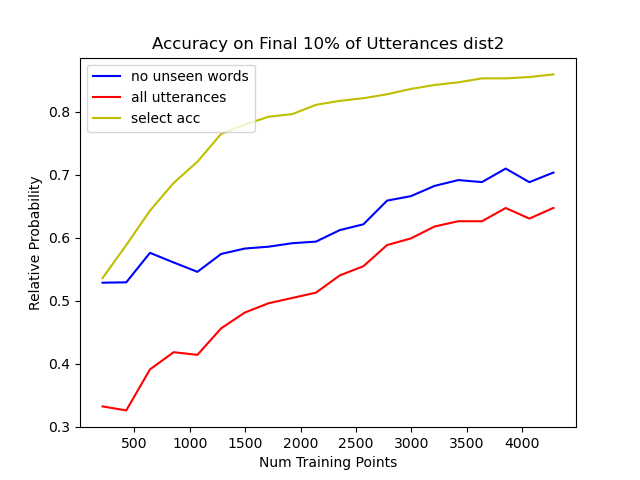}
    \includegraphics[width=0.45\linewidth]{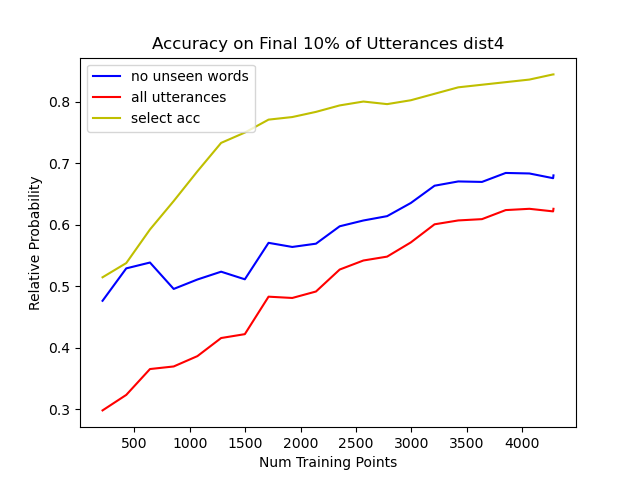}
    \includegraphics[width=0.45\linewidth]{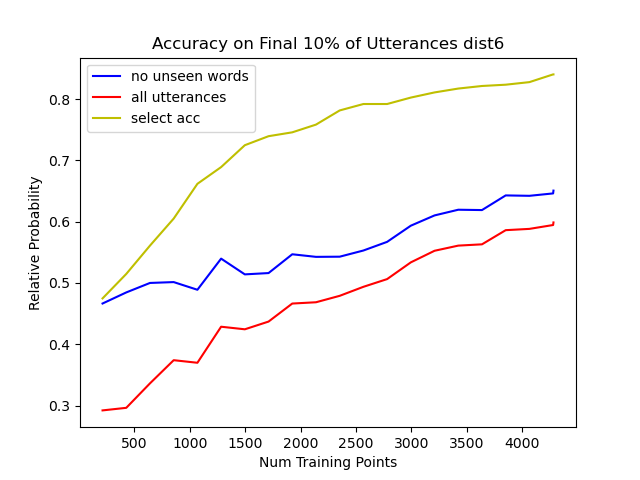}
    \includegraphics[width=0.45\linewidth]{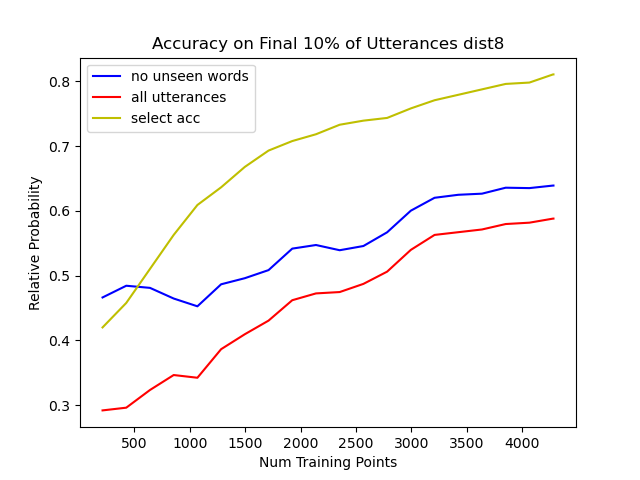}
    \caption{Repeats of experiments for predicting the correct utterance meaning at test time, with increasing numbers of distractor \acp{lf}: 2 in the top left, 4 in the top right, 6 in the bottom left and 8 in the bottom right. Cf. Figure \ref{fig:test-lfs}. \textcolor{black}{In this plot, and throughout the paper, a plot title ending in `distN' indicates that there were N distractors present. This shows that acquiring the ability to infer the meaning of whole utterances is robust to the presence of distractor meanings during training.}}
    \label{fig:test-lf-distractors}
\end{figure}

As shown in Figure \ref{fig:word-order-lex-entries-distractors}, the addition of distractors slows down learning, but the shape of the trajectories remains unchanged. This robustness represents an improvement on the model of \cite{abend2017bootstrapping}, which exhibited some instability with respect to the number of distractors. Figure \ref{fig:test-lf-distractors} shows the same stability for obtaining the correct meaning representation shown for the zero-distractor case in Figure 7. Here, the robustness to the distractor \acp{lf} is even more striking, showing only a very marginal difference even when 8 distractor \acp{lf} are added. This suggests that, with more training data, our model would reach the same performance as for the zero-distractor setting.

\FloatBarrier

\subsection{One-trial Learning of Nonce Words} \label{subsec:fast-mapping}
In this Section, we test the ability of our model for one-trial learning in a variety of syntactic contexts, that is, learning the meaning of novel words from a single exposure. 

\cite{abend2017bootstrapping} showed this in the case of transitive sentences. When exposed to a transitive sentence containing a nonce word `dax', along with two possible \acp{lf}, one in which `dax' means $\lambda x.\lambda y.dax\; x\; y$, and one in which it means $\lambda x.\lambda y.dax\; y\; x$, their model then showed a marked rise in its predicted probability that the meaning $\lambda x.\lambda y.dax\; x\; y$ is realized as the word `dax'. The most significant advance in our model over that of \citeauthor{abend2017bootstrapping} is its ability to handle a much wider set of syntactic constructions, and we now show that this allows our model to achieve one-trial learning over this wider set.

\begin{figure}
    \centering
    \includegraphics[width=0.45\linewidth]{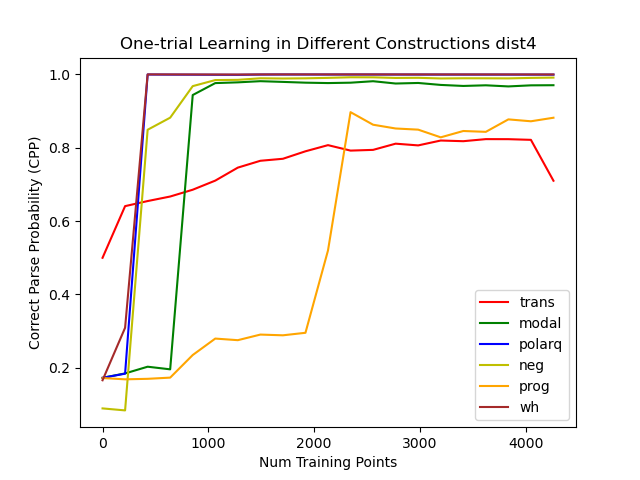}
    \includegraphics[width=0.45\linewidth]{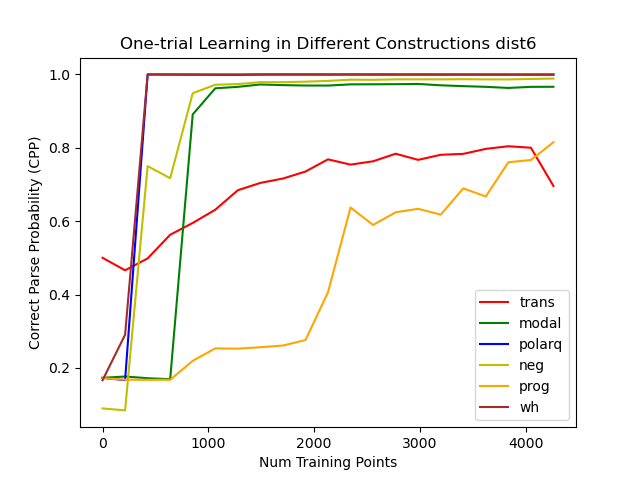}
    \caption{Evolution of the ability to learn from one-trial in the context of actor ambiguity. The y-axis shows the model`s estimated probability, after the single exposure, that the word ``dax" is assigned the syntactic type S $\bs$NP/NP and the logical form $\lambda\; x.\lambda\; y.dax\; x\; y$. \textcolor{black}{In this plot, and throughout the paper, a plot title ending in `distN' indicates that there were N distractors present. This shows that the model is able to acquire new word meanings from syntactic knowledge alone for a variety of construction types.}}
    \label{fig:one-trial-learning}
\end{figure}
Figure \ref{fig:one-trial-learning} shows the results of the same one-trial learning test not just for transitive sentences, but also for other more complex constructions that the child is exposed to. For each utterance type, the learner is presented two \acp{lf} that differ only in who they designate as the agent of the transitive verb, and only one, the intuitively correct one, agrees with the SVO verb category S$\bs$NP/NP. We follow \citeauthor{abend2017bootstrapping} in using two unseen names, `Jacob' and `Jacky' as subject and object, and in running two versions of the experiment, one with four distractor \acp{lf} and one with six. 
In the wh-question context, the utterance is ``who will Jacob dax?'', and one \ac{lf} expresses an object wh-question, while the other expresses a subject wh-question. 

Our learner, after training, is capable of one-trial learning in the context of all of these constructions. For questions, negations and modals, this measure of one-trial learning ability rises rapidly within the first 800 training examples. Such constructions contain some familiar words, namely the wh-word, the modal and the negation `not', so in this sense they are easier than the transitives, which contain only novel words. For the progressive `Jacob is daxing Jacky', the rise occurs later, slightly after training example 2000. The only familiar word there is the copula, which is in general a difficult word for the learner to analyse correctly because it appears frequently in a variety of different functional roles. The curve for transitives corresponds to curves presented in \citeauthor{abend2017bootstrapping}, and we can see that ours rise higher and more smoothly. 

This ability is the result of the model having enough language-specific syntax that, even if it encounters a new word, which must, in the first encounter, automatically get a low probability of having the correct meaning, the correct analysis has sufficiently high probability elsewhere in the tree to ensure the total tree probability is still high. This leads to a high update weight for the novel word being aligned with its correct meaning. 

The analysis in the earlier Figure \ref{fig:example-parse-tree} depicts an instance of this, of how already acquired lexical and syntactic knowledge can facilitate rapid acquisition of a new lexical item. There, the word ``music'' has never been observed before, and so the corresponding nodes of the tree have very low probability. However, by that stage, the learner is confident in the analysis of the rest of the sentence, so it still gives high probability to the depicted analysis. This high probability means that the co-occurrence counts for $music$ and ``music'' get a large update, leading to a large increase in the estimated probability that the former is the meaning of the latter. The difference with the experiments in this section is that the subject and object are also novel words, so the model must rely entirely on syntactic knowledge to determine how to relate the words to the components of the \ac{lf}.

The ability measured in Figure \ref{fig:one-trial-learning} is distinct from that of inferring the correct meaning for the utterance, as measured in Figure \ref{fig:parse-acc-by-type}. For example, if a verb and argument have been observed together several times, the learner may interpret the whole verb phrase as a single lexical item (an example of this is in Figure \ref{fig:prog-test-parses}b). This could give the correct meaning for the verb phrase, and hence the utterance, but, if such an analysis is made by the learner, it may not facilitate one-trial learning of a novel verb, because there it would not include a leaf node that contains just the novel verb word. For example, given the utterance ``he is daxing'', it could analyze ``is daxing'' as a single lexical item, at the expense of the analysis in which ``daxing'' is a leaf. So, although it might acquire the meaning for this entire VP in one trial, it would not do the same for the word ``daxing'' itself. 

Conversely, the learner may give the highest probability to an analysis that gives the \emph{incorrect} root \ac{lf} meaning, e.g. by interpreting ``is daxing'' as a single item meaning $\lambda x.\lambda y.see_{prog}\; x\; y$, while also having reasonably high probability on the correct analysis in which ``daxing'' is a root that means $\lambda x.\lambda y.dax_{prog}\; x\; y$. As ``daxing'' is not a leaf in the first analysis, its meaning distribution does not get any update, either correct or incorrect. Of the analyses in which it does get such an update, the correct meaning update may still dominate the probability mass, resulting in the model placing very high posterior probability on ``daxing'' having the correct meaning. In this case, the model would succeed at the one-trial learning test as measured in Figure \ref{fig:one-trial-learning}, but fail at the test of inferring utterance meaning, as measured in Figure \ref{fig:parse-acc-by-type}.

Figure \ref{fig:parse-acc-by-type} measures whether the single highest probability parse is correct, while \ref{fig:one-trial-learning} measures what fraction of the probability mass of the analyses in which the novel word is a leaf give it the correct meaning. The two measures give different, complementary views into the model`s learning trajectory across construction types. 

\subsection{Qualitative Results} \label{subsec:qual-examples}
\begin{figure}
    \centering
    $\colorbox{unseencolor}{not\; (will\; (hurt\; you\; it))}$
    \includegraphics[width=0.85\linewidth]{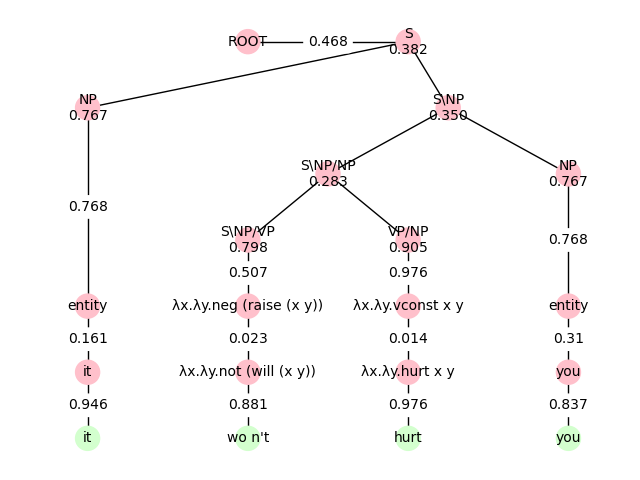}
    $\colorbox{unseencolor}{Q\; (shall\; (help\; you\; i))}$
    \includegraphics[width=0.85\linewidth]{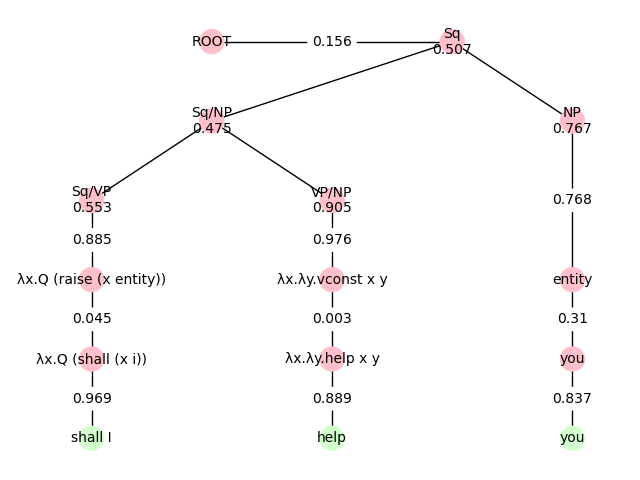}
    \caption{Examples of inferred parse trees for a negated (top) and an interrogative (bottom) modal utterance from our test set. The \ac{lf} shown above the trees are those inferred by the parse of the learner. Given information is in \colorbox{seencolor}{green}, inferred information is in \colorbox{unseencolor}{pink}. As this is test time, the model sees only the utterance and must infer the root \ac{lf}.}
    \label{fig:modal-test-parses}
\end{figure}

\begin{figure}
    \centering
    $\colorbox{unseencolor}{Q\; (do\; (need\; WHAT\; you))}$
    \includegraphics[width=0.85\linewidth]{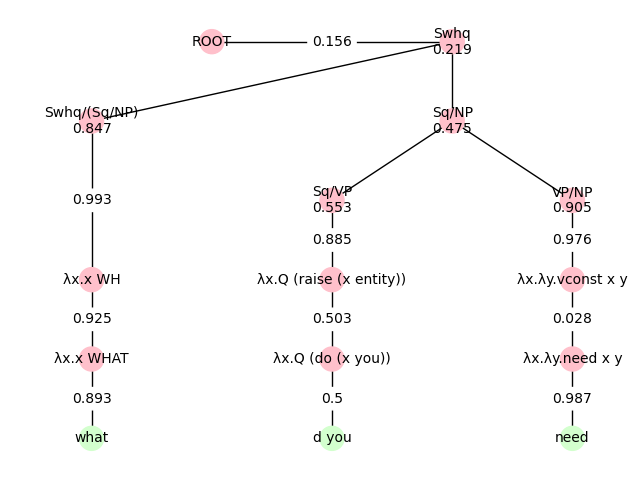}
    $\colorbox{unseencolor}{Q\; (does\; (say\; WHAT\; that))}$
    \includegraphics[width=0.85\linewidth]{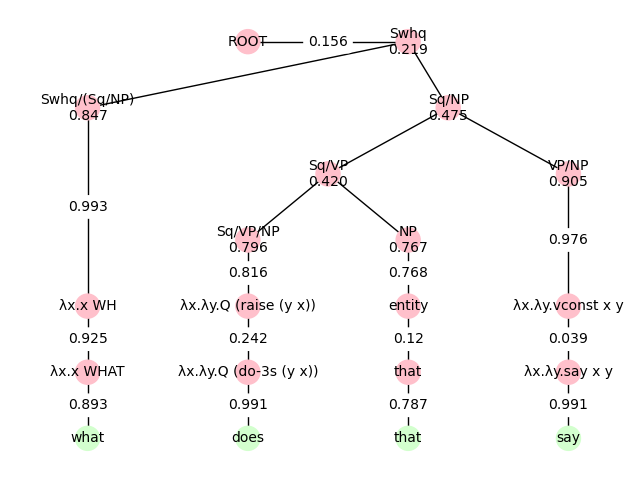}
    \caption{Examples of inferred parse trees for two object wh-questions from our test set. In the full theory, ``that'' would be raised as S/VP$\bs$(S/VP/NP). The \ac{lf} shown above the trees are those inferred by the parse of the learner. Given information is in \colorbox{seencolor}{green}, inferred information is in \colorbox{unseencolor}{pink}.}
    \label{fig:wh-test-parses}
\end{figure}

\begin{figure}
    \centering
    $\colorbox{unseencolor}{\ensuremath{pres_{3s}\; (check_{prog}\; (his\; watch)\; he)}}$
    \includegraphics[width=0.85\linewidth]{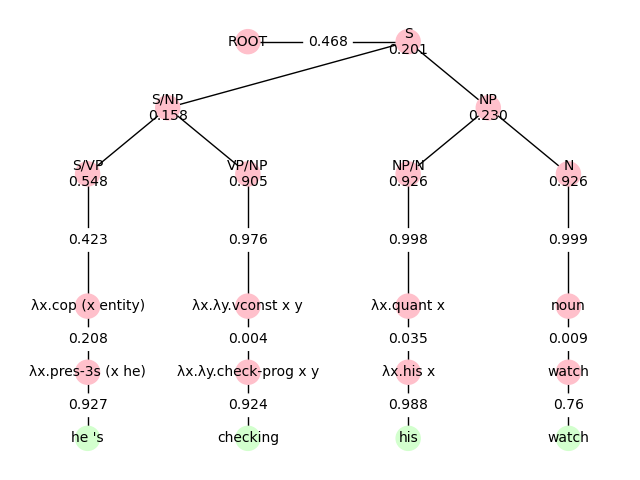}
    $\colorbox{unseencolor}{\ensuremath{Q\; (pres_{3s}\; (do_{prog}\; (WHAT)\; it))}}$
    \includegraphics[width=0.85\linewidth]{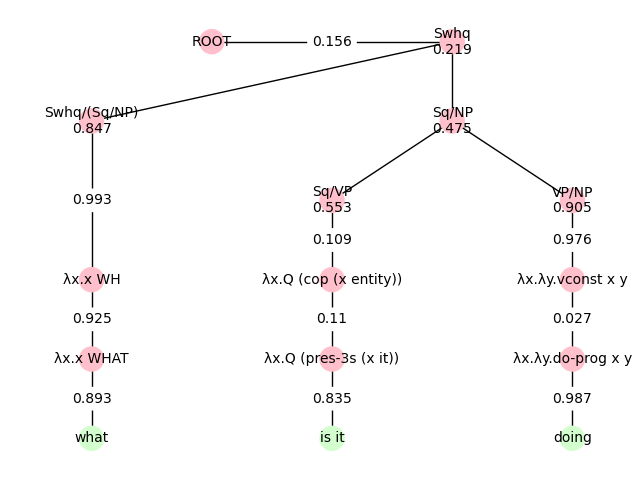}
    \caption{Example of inferred parse trees for progressives from our test set: one declarative (top) and one wh-question (bottom). The \ac{lf} shown above the trees are those inferred by the parse of the learner. Given information is in \colorbox{seencolor}{green}, inferred information is in \colorbox{unseencolor}{pink}.}
    \label{fig:prog-test-parses}
\end{figure}

Figures \ref{fig:modal-test-parses}, \ref{fig:wh-test-parses} and \ref{fig:prog-test-parses} show some examples from the final 10\% of utterances, which we use as a held-out test set. These examples come from presenting the model with the utterance only, and having it infer the parse tree and meaning, i.e., in Figure \ref{fig:test-lfs}, it corresponds to the red and blue lines, rather than the yellow line. 

We select examples that cover a range of the important constructions that our model is able to handle. Because of our special focus on \acp{lrd}, we show three object wh-questions, one in progressive aspect.

\begin{figure}
    \centering
    $\colorbox{unseencolor}{Q\; (not\; (see\; it\; you))}$
    \includegraphics[width=0.85\linewidth]{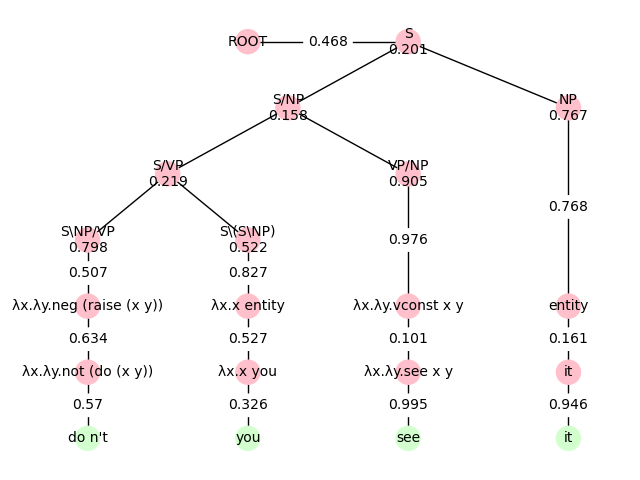}
    $\colorbox{unseencolor}{Q\; (will\; (give\; WHAT\; he\; you))}$
    \includegraphics[width=0.85\linewidth]{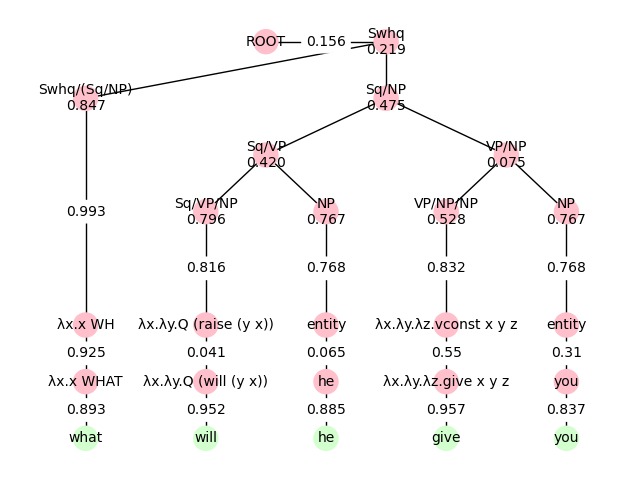}
    \caption{Example of an inferred parse tree for a polar negated question from our test set. Given information is in \colorbox{seencolor}{green}, inferred information is in \colorbox{unseencolor}{pink}.}
    \label{fig:hard-q-parses}
\end{figure}

In all of these examples, the learner infers a parse tree that will derive the correct root meaning. Some, such as the wh-question,``what does that say?'', in Figure \ref{fig:wh-test-parses}b, and the negated polar question in Figure \ref{fig:hard-q-parses}a, exhibit the standard, correct \ac{ccg} parse trees.
Others, such as Figure \ref{fig:modal-test-parses}a and \ref{fig:prog-test-parses}a, are still textbook-correct, though non-standard in the sense that they use composition when a purely applicative derivation is available. For the other two examples, the model interprets two orthographic words as a single lexical item, e.g. ``d you'', in the transcription of Figure \ref{fig:wh-test-parses}a, is interpreted as a single item with category S/VP and meaning $\lambda x.Q\; (do\; (x\; you))$. We observe that this often occurs for frequent bigrams. These examples still all end up with the correct root meaning for the whole utterance, so are all counted correct by the measure of Section \ref{subsec:meaning-results}. The tendency to lexicalize common multi-world expressions is discussed further in Section \ref{sec:discussion}.
Note that, in order to correctly analyze the wh-questions, the model has to select the correct question-form of the auxiliary ``do'', and the correct object-wh category for the wh-word ``what'', rather than the subject-wh category S$_{whq}$/VP or the in-situ category NP.

\FloatBarrier

\subsection{Summary of Empirical Improvements over Previous Models} \label{subsec:improvements-summary}
Compared to the two most similar existing models, \cite{abend2017bootstrapping} and  \cite{mahon2024language}, there are three main improvements offered by our model.
\begin{enumerate}
    \item It can handle a wider variety of syntactic constructions (Sections \ref{subsec:acc-by-const-type} and \ref{subsec:fast-mapping}), including long-range dependencies as found in object wh-questions (Section \ref{subsec:wh-results});
    \item It is the first to present evidence of fully correct parses for unseen utterances without corresponding \acp{lf} (Section \ref{subsec:qual-examples});
    \item It shows higher accuracy in inferring sentence meanings, and greater robustness to the inclusion of distractor \acp{lf} during training (Sections \ref{subsec:meaning-results} and \ref{subsec:distractor-results}).
\end{enumerate}

\section{Discussion} \label{sec:discussion}
\subsection{Long-range Dependencies}
It is clear from both the breakdown of accuracy by construction type, as shown in Figure \ref{fig:parse-acc-by-type}, and from the qualitative examples in Section \ref{subsec:qual-examples}, that the model largely succeeds in learning the long-range dependencies in object wh-questions. In fact, in Figure \ref{fig:parse-acc-by-type}, the learning curve for wh-questions (pink line) is, for most of training, the highest of all construction types. However, this precocity is due to the fact that there are a couple of frequent wh question utterances, such as ``what are you doing?'' and ``what`s that?'', that the model learns to memorize very early, which accounts for the early jump. By `memorize', we mean that the entire utterance is modelled as a single word, as discussed in detail in Section \ref{subsec:ngram-lexicalisation}. The subsequent gradual rise of the pink line is then caused by the model learning the general form of wh-questions and getting more of the long tail correct. By the end of training, it can correctly analyse the large majority of novel wh-questions, producing the textbook \ac{ccg} parse tree as well and meaning, even including some ditransitives, such as ``what will he give you?'', in Figure \ref{fig:hard-q-parses}b. 



\subsection{Lexicalization of Common Ngrams} \label{subsec:ngram-lexicalisation}
A common behaviour of our learner is to treat multiple orthographic words as a single lexical item, e.g. ``is it'' in Figure \ref{fig:prog-test-parses}b. It is important to allow this interpretation, rather than tell the learner explicitly where the word boundaries are, because we assume that, while learned phonotactic constraints are able to identify possible word boundaries, they are not able to determine them exactly, and so the child, when learning syntax and semantics, must also be considering such interpretations. In our data, the possible boundaries are those provided by the UD tokenizer, as used in \cite{ida2023}. This amounts to potential boundaries at all spaces and around clitics. 

As our model is entirely probabilistic, there is no hard line between being in or out of the lexicon: every n-gram that was observed anywhere during training has some probability of being the word for the corresponding \ac{lf}, but for most n-grams, this probability is negligible, and it will never appear in the single \ac{map} analysis. In this discussion, we use the term `lexicalisation' to refer to the case where an ngram appears with significant frequency in the \ac{map} analysis.

In the extreme case, the learner might ignore all potential boundaries and treat the entire utterance as a single word. Indeed, the following simple probabilistic analysis shows that, in our framework, this is the null hypothesis in that, prior to seeing any training data, it is the favoured analysis of all utterances. The probability of a parse tree is the product of the probabilities of all of the constituent nodes given their parents. Prior to seeing any training data, the probability of a category being a leaf is the same for all categories, and is strictly less than 1 (because some probability must be reserved for other possible splits of that category). Thus, the probability of the parse tree is minimized when there is just a single leaf. However, as training progresses, the model favours larger and larger parse trees and eventually, in many cases, reaches a stable interpretation comprising the standard \ac{ccg} parse tree. The moment when this point is reached is the moment at which one-trial learning, as in Section \ref{fig:one-trial-learning}, becomes possible. 

Mostly, these cases correspond to breaking at all word boundaries, but there are some exceptions. The negation contraction ``n't'' is almost always analysed as a single lexical item together with the auxiliary, as in Figures \ref{fig:modal-test-parses} and \ref{fig:hard-q-parses}a, even though there is a potential word boundary between them. This in fact agrees with standard linguistic assumptions \citep{Bybee:02a}, and there is strong evidence that negated auxiliary contractions are single items in adult lexicons, as they can be inverted, while the un-contracted bigram cannot: \textit{don't you see it} vs *\textit{do not you see it}. Many common bigrams that are lexicalised by our model agree with contractions in adult speech, e.g. ``d'you'' in Figure \ref{fig:wh-test-parses}a and ``he's'' in Figure \ref{fig:prog-test-parses}a\footnote{Indeed, one could make a case either for or against including a potential boundary for words that were transcribed as clitics. The reason we do is simply that the universal dependency parser used by \cite{ida2023}, and hence the data we use, does so.}, but there are also several that do not: ``'is it'' in Figure \ref{fig:prog-test-parses}b and in Figure \ref{fig:hard-q-parses}b. For some, such as ``that's right'' and ``I don't know'', the \ac{map} analysis remains as a one-word utterance even at the end of training. This is also consistent with adult spoken contractions transcribed as ``s'right'' and ``I`d`no''.

Discussions of lexicalisation have identified several aspects to the process \citep{bauer1983english}, e.g. prosodic lexicalization--the effect on the phonetic realization of the segment of the utterance--morphological lexicalisation--the characterisation of irregular inflected forms as being lexicalised, semantic lexicalization, a.k.a. idiomatization \citep{lipka1977lexikalisierung}, and the effect of frequency on lexicalisation \citep{Langacker:88a,Lieven:03a,Bybee:06,Bannard:08a}. 

Of the possible causes of lexicalisation, the only one that our model responds to is frequency. If the frequency of an ngram is large enough that the probability of it being paired with its corresponding meaning is greater than the product of the probabilities of each its constituents being paired with their corresponding meanings, then it will be lexicalised, in the sense outlined above. 
The closest analogue in human lexicalisation, would be to consider as lexicalised so-called conventionalised colocations, a.k.a ``prefabs'' \citep{erman2000idiom}--that is, ngrams that are not idiomatic but appear unusually frequently, such as `ulterior motive'. Such a picture has been suggested by \cite{Bybee:06}, \cite{erman2000idiom} and \cite{bybee1985morphology}. Note that what counts, at least in the case of our model, is not the raw occurrence frequency of the ngrams in the corpus, 
but rather the frequency of the ngrams in the estimated parse trees. This difference means that ngrams that cohere with the rest of the sentence into a probable parse tree count for more than those that do not. These observations may also support a lexical analysis of processes of cliticization.


The tendency of our model to lexicalise certain ngrams suggests that, from a probabilistic model of syntax and semantics alone, there is a signal to do so. However, without the other components such as phonetics, the choice for such lexicalizations may differ somewhat from those evidenced in humans.

\subsection{Modeling Morphology and Phonology}
The potential future extensions to include morphology were also discussed in the model of \cite{mahon2024language}, which is similar to ours. \citet{mahon2024language} outlined two possible approaches to including morphology: either to extend the \ac{ccg} parse trees down to the level of morphemes, or else replace the Dirichlet process for predicting word form given meaning with a neural model. We explored the former idea in preliminary experiments and found it not to work well. For example, we tried inserting a potential word boundary between  the suffix `-ing' and the root, in the idea that it model could learn `-ing' had the category VP$\bs$(S$\bs$NP) (for transitive sentences) and meaning $\lambda p.\lambda x.prog(p\; x)$. However, it always chose to interpret the stem and the suffix together as a single lexical item. The second idea, of using a neural predictor of word form, could be more promising, as it may allow the model to learn some systematic relationship between meaning and word form without having to specify something as precise as that the orthographic suffix `-ing' always indicates progressive aspect. This neural predictor could operate on the IPA transcriptions instead of the orthographic ones 
, or even on the speech waveform itself, both of which are available in the CHILDES corpus we use. This would be to take a position that the syntax-semantics interface can be learnt in part by a symbolic system (namely, the one we present in the present paper), but that morphology is more suited to a connectionist model, which is consistent with the success of finite-state transducers in morphological analysis \citep{Kay:87a}. In Section 3.1, we noted that such a mechanism is expected to be needed, in the form of the probabilistic supertagger, for the resolution of lexical ambiguity as the grammar grow towards adult size, and this would constitute a parallel "thinking fast" grammatical component to to symbolic "thinking slow" \citep{Kahneman:17a, Ferr:07}. \textcolor{black}{\cite{Stanojevic:23a} offer neuropsychological veviddence for the
involvement of such a hybrid symbolic-neurocomputational mechanism in
human sentence processing using a fully incremental parsing algorithm
combined with a supertagger.Providing such a morphological analyser will be a necessary first step in demonstrating the universality of our syntactic learner by applying it to the similarly annotated Hagar corpus of Hebrew child-directed utterance described by \citeauthor{ida2023}.}



\section{Conclusion}
This paper presented a computational model for child language acquisition of syntax and word-level semantics, trained on transcribed child-directed speech paired with manually annotated logical forms as meaning representations. 
Our model works with several orders of magnitude less data than even the most sample-efficient transformer-based approaches to modelling human-like learning of language \citep{warstadt2023findings}. The main advances of our model over previous similar ones lie in increased robustness and stability in learning, the extension to a wider range of constructions, and the ability to infer meanings and parse trees for unseen child-directed utterance from the held-out final sample of the corpus. We replicated the experiments of previous similar models regarding learning word order and word meanings, and showed that our model has 80\% accuracy on inferring the meaning of novel utterances. While prior works demonstrated some limited ability for one-trial learning of word meanings in simple transitive sentences, our model learns these word meanings very rapidly and confidently in a wide variety of construction types. Finally, we discussed the model`s handling of long-range dependencies, and its tendency to lexicalize common ngrams and how this might relate to usage-based lexicalization in humans. Despite the comparatively impoverished nature of our training datasets, the model's ability to acquire constructions, including those involving long-range dependencies, and its tendencies both to lexicalize frequent collocations and later to re-analyse them compositionally, appear to be broadly consistent with the course of language acquisition in real children.

\section{Acknowledgements}
This research was supported by ERC Advanced Fellowship GA 742137
SEMANTAX and the University of Edinburgh Huawei Laboratory.

\bibliographystyle{elsarticle-harv} 
\bibliography{bibliography}

\appendix 

\section{Mapping from CHILDES POS Tags to Montagovian Semantic Types} \label{app:childes-to-semcats}
Table~\ref{tab:mapping-childes-to-semcat} shows how we infer the Montagovian semantic type from the CHILDES POS tags that are available in our \acp{lf}. Some are defined schematically, the avoid overly long expressions. For example, the category for conjunctions (conj) and coordinations (coord), we use the variable $X$ to stand for any other semantic category. The reason the mapping from tags to semantic types is many-to-one is that this allows learning to be shared across categories. For example, if the model learns that the general category ‘det’ precedes nouns, it knows that this is true for all types of determiners, whereas if we distinguish between ‘det:art’, ‘det:poss’, ‘det:num’ etc., then it has to learn this separately for each.

\begin{table} 
\caption{Our mapping from CHILDES part of speech tags of terms in the logical form to Montagovian semantic types.} \label{tab:mapping-childes-to-semcat}
\centering
\resizebox{!}{0.6\textwidth}{
\begin{tabular}{ll}
\toprule
CHILDES TAG & const marking in shell \ac{lf} \\
\midrule
adj & \textless \textless e,t\textgreater ,\textless e,t\textgreater \textgreater \\
adv & not considered \\
adv:int & not considered \\
adv:tem & not considered \\
aux & not considered \\
conj & \textless X,\textless X,X\textgreater \textgreater  \\
coord & \textless X,\textless X,X\textgreater \textgreater \\
cop & handled separately \\
det & \textless \textless e,t\textgreater ,e\textgreater  \\
det:art & \textless \textless e,t\textgreater ,e\textgreater  \\
det:dem & \textless \textless e,t\textgreater ,e\textgreater  \\
det:int & \textless \textless e,t\textgreater ,e\textgreater  \\
det:num & \textless \textless e,t\textgreater ,e\textgreater  \\
det:poss & \textless \textless e,t\textgreater ,e\textgreater  \\
mod & \textless \textless \textless e,t\textgreater ,\textless e,t\textgreater \textgreater ,\textless e,t\textgreater \textgreater  \\
mod:aux & \textless \textless e,t\textgreater ,e\textgreater  \\
n & \textless e,t\textgreater  \\
n:pt & \textless e,t\textgreater  \\
n:gerund & e \\
n:let & e \\
n:prop & e \\
neg & \textless \textless e,\textless e,t\textgreater \textgreater ,\textless e,\textless e,t\textgreater \textgreater \textgreater ,\;  \textless \textless e,t\textgreater ,\textless e,t\textgreater \textgreater   {t,t} \\
prep & \textless \textless e,t\textgreater, \textless e, t\textgreater \textgreater \\
pro:dem & e \\
pro:indef & e \\
pro:int & e \\
pro:obj & e \\
pro:per & e \\
pro:poss & \textless e,t\textgreater  \\
pro:refl & e \\
pro:sub & e \\
qn & \textless e,t\textgreater  \\
v & \textless e,\textless e,t\textgreater \textgreater ,\; \textless e,t\textgreater  \\
\bottomrule
\end{tabular}
}
\end{table}

\subsection{Manually Annotated Lexicon for Fifty Most Common Words} \label{app:annotated-lexicon}
This section shows the ground-truth logical form meaning representation and \ac{ccg} syntactic category for the fifty most common words in each dataset. As described in Section~\ref{subsec:single-word-results}, these are used to evaluate the learner's ability to acquire the correct lexicon. Note, the \acp{lf} that appeared in the main paper were abbreviated for clarity. Here, we write the full \ac{lf}, including the CHILDES part of speech tag. The full lexical entry is of the form \text{\textlangle\ac{lf}\textrangle} $|\, |$ \text{\textlangle syntactic-category\textrangle}

. Where a word has two common meanings, we include two different lexical entries, separated with a comma.
\vspace{2ex}

\footnotesize
\begin{lstlisting}
'll:$\lambda$ x.$\lambda$ y.mod|~will (x y) || S\\NP/(S\\NP)
're:$\lambda$ x.$\lambda$ y.v|hasproperty y x || S\\NP/NP,
    $\lambda$ x.$\lambda$ y.v|equals y x || S\\NP/NP
's:$\lambda$ x.$\lambda$ y.v|equals y x || S\\NP/NP,
    $\lambda$ x.$\lambda$ y.v|hasproperty y x || S\\NP/NP
Adam:n:prop|adam || NP
I:pro:sub|i || NP
a:$\lambda$ x.det:art|a x || NP/N
an:$\lambda$ x.det:art|a x || NP/N
another:$\lambda$ x.qn|another x || NP/N
are:$\lambda$ x.$\lambda$ y.v|equals x y || S\\NP/NP,
    $\lambda$ x.$\lambda$ y.v|hasproperty y x || S\\NP/NP
break:$\lambda$ x.$\lambda$ y.v|break y x || S\\NP/NP
can:$\lambda$ x.$\lambda$ y.mod|can (x y) || S\\NP/(S\\NP),
    $\lambda$ x.$\lambda$ y.mod|can (x y) || S/NP/(S\\NP)
d:$\lambda$ x.$\lambda$ y.mod|do (x y) || S\\NP/(S\\NP),
    $\lambda$ x.$\lambda$ y.mod|do (x y) || S/NP/(S\\NP)
did:$\lambda$ x.$\lambda$ y.mod|do-past (x y) || S/NP,
    $\lambda$ x.$\lambda$ y.mod|do-past (x y) || S/NP/(S\\NP)
do:$\lambda$ x.$\lambda$ y.v|do y x || S\\NP/NP,
    $\lambda$ x.$\lambda$ y.mod|do (x y) || S/NP/(S\\NP)
does:$\lambda$ x.$\lambda$ y.mod|do-3s (y x) || S\\NP/(S\\NP),
    $\lambda$ x.$\lambda$ y.mod|do-3s (x y) || S/NP/(S\\NP)
dropped:$\lambda$ x.$\lambda$ y.v|drop-past y x || S\\NP/NP
have:$\lambda$ x.$\lambda$ y.v|have y x || S\\NP/NP
he:pro:sub|he || NP
his:$\lambda$ x.det:poss|his x || NP/N,pro:poss|his || NP
hurt:$\lambda$ x.$\lambda$ y.v|hurt-zero y x || S\\NP/NP
in:$\lambda$ x.$\lambda$ y.prep|in (y x) || S\\NP\\(S\\NP)/NP,
    $\lambda$ x.prep|in x || S/S
is:$\lambda$ x.$\lambda$ y.v|equals x y || S\\NP/NP,
    $\lambda$ x.$\lambda$ y.v|hasproperty y x || S\\NP/NP
it:pro:per|it || NP
like:$\lambda$ x.$\lambda$ y.v|like y x || S\\NP/NP
lost:$\lambda$ x.$\lambda$ y.v|lose-past y x || S\\NP/NP
may:$\lambda$ x.$\lambda$ y.mod|may (x y) || S\\NP/(S\\NP)
missed:$\lambda$ x.v|miss-past x || S\\NP,
    $\lambda$ x.$\lambda$ y.v|miss-past y x || S\\NP/NP
my:$\lambda$ x.det:poss|my x || NP/N
name:n|name || N
need:$\lambda$ x.$\lambda$ y.v|need y x || S\\NP/NP
no:$\lambda$ x.qn|no x || NP/N
not:$\lambda$ x.$\lambda$ y.not (x y) || S\\NP/(S\\NP)\(S\\NP/(S\\NP))
on:$\lambda$ x.prep|on x || S\\NP\\(S\\NP)/NP
one:pro:indef|one || NP
pencil:n|pencil || N
say:$\lambda$ x.$\lambda$ y.v|say y x || S\\NP/NP
see:$\lambda$ x.$\lambda$ y.v|see y x || S\\NP/NP
shall:$\lambda$ x.$\lambda$ y.mod|shall (x y) || S\\NP/(S\\NP)
some:$\lambda$ x.qn|some x || NP/N
that:pro:dem|that || NP,$\lambda$ x.pro:det|that x || NP/N
the:$\lambda$ x.det:art|the x || NP/N
they:pro:sub|they || NP
this:pro:dem|this || NP,$\lambda$ x.pro:det|this x || NP/N
those:pro:dem|those || NP,$\lambda$ x.pro:det|those x || NP/N
was:$\lambda$ x.$\lambda$ y.v|equals x y || S\\NP/NP,
    $\lambda$ x.$\lambda$ y.v|hasproperty y x || S\\NP/NP
we:pro:sub|we || NP
what:pro:int|WHAT || Swhq/Sq/NP,pro:int|WHAT || NP
who:pro:int|WHO || Swhq/Sq/NP,pro:int|WHO || NP
you:pro:per|you || NP
your:$\lambda$ x.det:poss|your x || NP/N

\end{lstlisting}

\section{Mapping from CHILDES POS Tags to Shell \ac{lf} Terms}
As described in Section~\ref{subsec:prob-model}, we use the CHILDES part of speech tags, which are included in the logical forms of \cite{ida2023}, to choose the marking on the constant in the shell logical form. Table~\ref{tab:conversion-chiles-shell} gives full correspondence. In the main text in Section~\ref{subsec:word-order}, we indicated the marking with the first letter of the right column, e.g. `verb' gives `vconst'.

\begin{table} 
\caption{Our mapping from CHILDES part of speech tags of terms in the logical form to the marking on the constant in the corresponding shell logical form.} \label{tab:conversion-chiles-shell}
\centering
\resizebox{!}{0.6\textwidth}{
\begin{tabular}{ll}
\toprule
CHILDES TAG & const marking in shell \ac{lf} \\
\midrule
adj & adj \\
adv & adv \\
adv:int & adv \\
adv:tem & adv \\
aux & aux \\
conj & connect \\
coord & connect \\
cop & cop \\
det & quant \\
det:art & quant \\
det:dem & quant \\
det:int & quant \\
det:num & quant \\
det:poss & quant \\
mod & raise \\
mod:aux & quant \\
n & noun \\
n:pt & noun \\
n:gerund & entity \\
n:let & entity \\
n:prop & entity \\
neg & neg \\
prep & prep \\
pro:dem & entity \\
pro:indef & entity \\
pro:int & WH \\
pro:obj & entity \\
pro:per & entity \\
pro:poss & quant \\
pro:refl & entity \\
pro:sub & entity \\
qn & quant \\
v & verb \\
\bottomrule
\end{tabular}
}
\end{table}

\section{Base Distributions} \label{app:base-distributions}
As described in Section \ref{subsec:prob-model}, each of the components of our model, $p_r$, $p_t$, $p_e$, $p_l$ and $p_w$ use a base distribution, which is then updated with the expected observed cooccurrence counts during training. The base distributions for each of these models are as follows:
\begin{itemize}
    \item for $p_r$ and $p_t$: $H(y) = 0.9^{n}$, where $n$ is the number of atomic categories in $y$;
    \item for $p_l$ and $p_e$, $H(y) = 0.25^{n}$, where $n$ is the number of variables and constants in $y$;
    \item for $p_w$, $H(y) = 0.72^n$, where $n$ is the number of letters in $y$.
\end{itemize}
These are unnormalised distributions, because they do not sum to 1, though their sum is finite. In principle, these could be normalised by fixing a vocabulary size, however we simply leave them unnormlised in our experiments. Normalisation is immaterial for $p_w$ anyway, because all analyses for an observed utterance will have the same number of letters in the leaf node words, so normalising would just multiply the numerator and the denominator of Equation \eqref{eq:bayes-posterior} by the same factor.

\end{document}